\documentclass[sigconf,nonacm]{acmart}

\usepackage{times}
\usepackage{latexsym}
\usepackage{textgreek}
\usepackage[utf8]{inputenc} % 如果使用 PDFLaTeX
\usepackage{makecell}
\usepackage{tabularx}
\usepackage{array} % 支持 p{width} 形式的列定义
\usepackage{multirow}
\usepackage{changepage}
\usepackage[T1]{fontenc}
\usepackage[utf8]{inputenc}
\usepackage{alphabeta}
\usepackage{threeparttable}
\usepackage{microtype}

\usepackage{graphicx}
\usepackage{booktabs}
\usepackage{amsmath}
\usepackage{tcolorbox}
\usepackage{verbatim}
\AtBeginDocument{%
  }

\setcopyright{acmlicensed}
\copyrightyear{2018}
\acmYear{2018}
\acmDOI{XXXXXXX.XXXXXXX}
\acmConference[Conference acronym 'XX]{Make sure to enter the correct
  conference title from your rights confirmation email}{June 03--05,
  2018}{Woodstock, NY}
\acmISBN{978-1-4503-XXXX-X/2018/06}

\acmSubmissionID{123-A56-BU3}

\begin{document}

%%
%% The "title" command has an optional parameter,
%% allowing the author to define a "short title" to be used in page headers.
\title[Plutus]{Plutus: Benchmarking Large Language Models in Low-Resource Greek Finance}
% \subtitle{Astylo Meets Plutus}

%%
%% The "author" command and its associated commands are used to define
%% the authors and their affiliations.
%% Of note is the shared affiliation of the first two authors, and the
%% "authornote" and "authornotemark" commands
%% used to denote shared contribution to the research.
% \author{Ben Trovato}
% \authornote{Both authors contributed equally to this research.}
% \email{trovato@corporation.com}
% \orcid{1234-5678-9012}
% \author{G.K.M. Tobin}
% \authornotemark[1]
% \email{webmaster@marysville-ohio.com}
% \affiliation{%
%   \institution{Institute for Clarity in Documentation}
%   \city{Dublin}
%   \state{Ohio}
%   \country{USA}
% }

\author{\small Xueqing Peng}
\affiliation{%
\small
  \institution{The Fin AI}
  \country{USA}}
\email{xueqing.peng2023@gmail.com}

\author{\small Triantafillos Papadopoulos}
% \author{\small Efstathia Soufleri}
\affiliation{%
\small
  \institution{Athens University of Economics and Business, Archimedes/Athena RC}
  % \institution{ Archimedes/Athena RC}
  \city{Athens}
  \country{Greece}}
\email{t.papadopoulos@athenarc.gr}
% \email{{t.papadopoulos,e.soufleri}@athenarc.gr}

\author{\small Efstathia Soufleri}
\affiliation{%
\small
  \institution{Archimedes/Athena RC}
  \city{Athens}
  \country{Greece}}
\email{e.soufleri@athenarc.gr}

\author{\small Polydoros Giannouris}
\affiliation{%
\small
  \institution{The University of Manchester}
  \city{Manchester}
  \country{UK}}
\email{polydoros.giannouris@postgrad.manchester.ac.uk}

\author{\small Ruoyu Xiang}
% \author{\small Yan Wang}
% \author{\small Lingfei Qian}
\affiliation{%
\small
  \institution{The Fin AI}
  \country{USA}
}
\email{xry0408@gamail.com}
% \email{{xry0408,wy2266336,lfqian94}@gmail.com}

\author{\small Yan Wang}
\affiliation{%
\small
  \institution{The Fin AI}
  \country{USA}
}
\email{wy2266336@gmail.com}

\author{\small Lingfei Qian}
\affiliation{%
\small
  \institution{The Fin AI}
  \country{USA}
}
\email{lfqian94@gmail.com}

\author{\small Jimin Huang}
% \author{\small Qianqian Xie}
\affiliation{%
\small
  \institution{The Fin AI}
  \country{USA}
}
\email{jimin.huang@thefin.ai}
% \email{{jimin.huang@thefin.ai,xqq.sincere@gmail.com}}

\author{\small Qianqian Xie}
\affiliation{%
\small
  \institution{The Fin AI}
  \country{USA}
}
\email{xqq.sincere@gmail.com}

\author{\small Sophia Ananiadou}
\affiliation{%
\small
  \institution{The University of Manchester}
  \city{Manchester}
  \country{UK}
}
\affiliation{%
\small
  \institution{Archimedes/Athena RC}
  \city{Athens}
  \country{Greece}}
\email{sophia.ananiadou@manchester.ac.uk}

%%
%% By default, the full list of authors will be used in the page
%% headers. Often, this list is too long, and will overlap
%% other information printed in the page headers. This command allows
%% the author to define a more concise list
%% of authors' names for this purpose.
\renewcommand{\shortauthors}{Xueqing et al.}

%%
%% The abstract is a short summary of the work to be presented in the
%% article.
\begin{abstract}
Despite Greece’s pivotal role in the global economy, large language models (LLMs) remain underexplored for Greek financial context due to the linguistic complexity of Greek and the scarcity of domain-specific datasets. Previous efforts in multilingual financial natural language processing (NLP) have exposed considerable performance disparities, yet no dedicated Greek financial benchmarks or Greek-specific financial LLMs have been developed until now. To bridge this gap, we introduce \textbf{Plutus-ben}, the first Greek Financial Evaluation Benchmark, and \textbf{Plutus-8B}, the pioneering Greek Financial LLM, fine-tuned with Greek domain-specific data. Plutus-ben addresses five core financial NLP tasks in Greek: numeric and textual named entity recognition, question answering, abstractive summarization, and topic classification, thereby facilitating systematic and reproducible LLM assessments. To underpin these tasks, we present three novel, high-quality Greek financial datasets, thoroughly annotated by expert native Greek speakers, augmented by two existing resources. Our comprehensive evaluation of 22 LLMs on Plutus-ben reveals that Greek financial NLP remains challenging due to linguistic complexity, domain-specific terminology, and financial reasoning gaps. These findings underscore the limitations of cross-lingual transfer, the necessity for financial expertise in Greek-trained models, and the challenges of adapting financial LLMs to Greek text. We release Plutus-ben, Plutus-8B, and all associated datasets\footnote{~\url{https://huggingface.co/collections/TheFinAI/plutus-benchmarking-greek-financial-llms-67bc718fb8d897c65f1e87db}} publicly to promote reproducible research and advance Greek financial NLP, fostering broader multilingual inclusivity in finance.
\end{abstract}

%%
%% The code below is generated by the tool at http://dl.acm.org/ccs.cfm.
%% Please copy and paste the code instead of the example below.
%%
\begin{CCSXML}
<ccs2012>
   <concept>
       <concept_id>10010147.10010178.10010179.10010186</concept_id>
       <concept_desc>Computing methodologies~Language resources</concept_desc>
       <concept_significance>500</concept_significance>
       </concept>
 </ccs2012>

 <ccs2012>
   <concept>
       <concept_id>10010405.10010497.10010510.10010513</concept_id>
       <concept_desc>Applied computing~Annotation</concept_desc>
       <concept_significance>500</concept_significance>
       </concept>
 </ccs2012>
\end{CCSXML}

\ccsdesc[500]{Applied computing~Annotation}
\ccsdesc[500]{Computing methodologies~Language resources}

%%
%% Keywords. The author(s) should pick words that accurately describe
%% the work being presented. Separate the keywords with commas.
\keywords{Financial Large Language Models, Financial Application, Benchmark, Greek, Low-resource Languages}
%% A "teaser" image appears between the author and affiliation
%% information and the body of the document, and typically spans the
%% page.

% \begin{teaserfigure}
%   \includegraphics[width=\textwidth]{sampleteaser}
%   \caption{Seattle Mariners at Spring Training, 2010.}
%   \Description{Enjoying the baseball game from the third-base
%   seats. Ichiro Suzuki preparing to bat.}
%   \label{fig:teaser}
% \end{teaserfigure}

\received{20 February 2007}
\received[revised]{12 March 2009}
\received[accepted]{5 June 2009}

%%
%% This command processes the author and affiliation and title
%% information and builds the first part of the formatted document.
\maketitle

\begin{figure}[h]
  \centering
  \includegraphics[width=\linewidth]{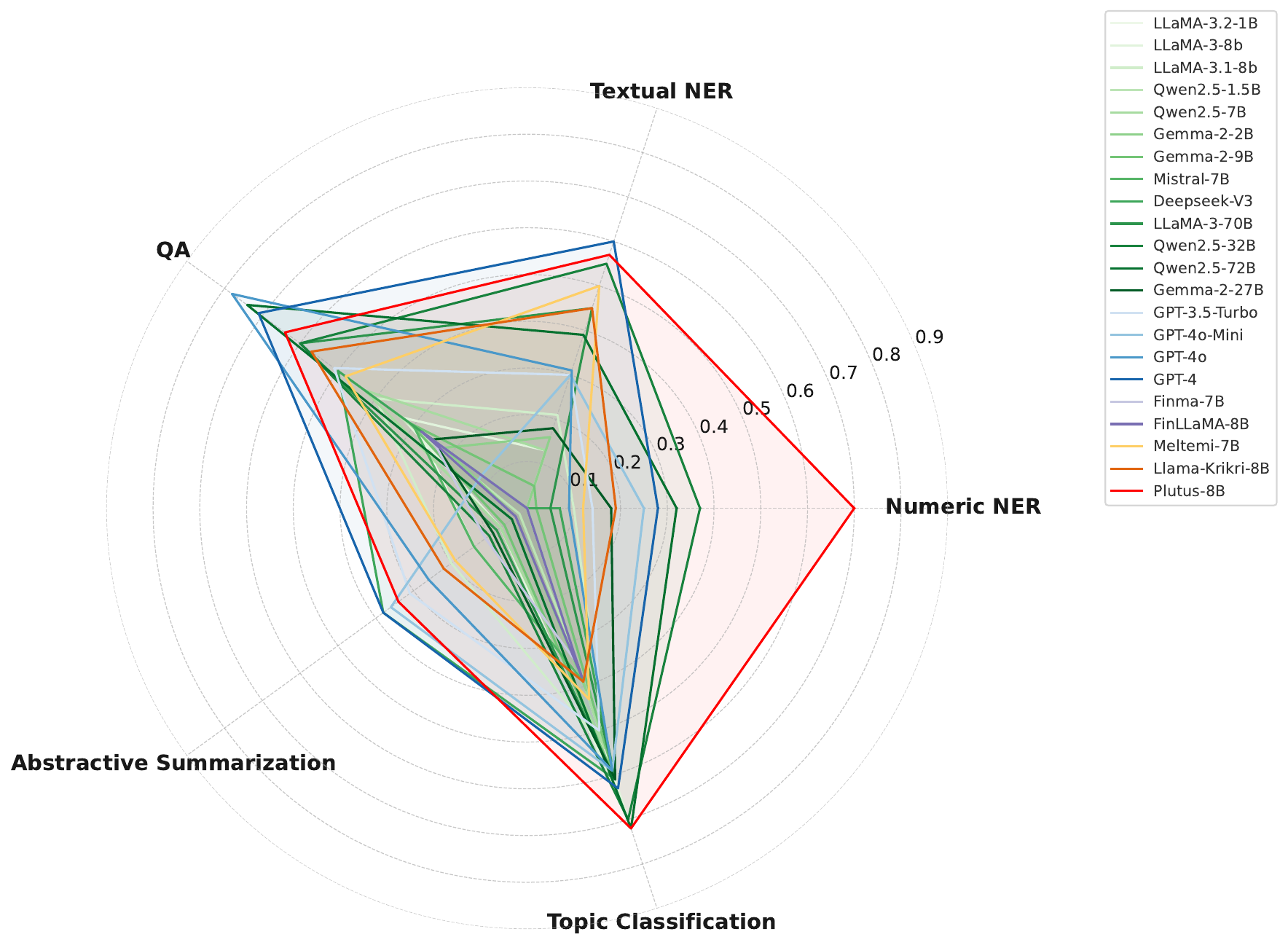}
  \caption{Radar graph of model performance on Plutus-ben, the first Greek financial benchmark. Plutus-8B achieves the best performance, surpassing GPT-4 by 15.38\%, GPT-4o by 46.34\%, and Deepseek-V3 by 93.55\%.}
  % \Description{A woman and a girl in white dresses sit in an open car.}
    \label{fig:radar}
\end{figure}

\section{Introduction}
As an official language of the European Union\footnote{~\url{https://european-union.europa.eu/principles-countries-history/languages_en}} and the dominant language of Greece’s merchant navy, which controls over 20\% of the world’s merchant fleet\footnote{~\url{https://ugs.gr/en/greek-shipping-and-economy/greek-shipping-and-economy-2024/the-international-perspective/}}, Greek is central to international trade, banking, and regulatory affairs. Greek financial documents such as regulatory filings, maritime trade records, and economic reports hold substantial international relevance, yet their processing remains difficult~\cite{esarey2020lessons}. Greek’s complex morphology, inflectional system, and unique orthographic structures~\cite{holton2012greek, efthymiouinflectional} make it fundamentally different from high-resource financial languages such as English and Chinese. These linguistic complexities introduce challenges in financial information extraction, entity recognition, and numerical reasoning~\cite{papantoniou2024nlpgreeklanguagelonger}.

Despite recent advancements in applying large language models (LLMs) to financial natural language processing (NLP) tasks, Greek remains largely unexplored. Extensive financial LLMs have been developed for English~\citep{xie2024openfinllmsopenmultimodallarge, DBLP:journals/corr/abs-2303-17564, DBLP:conf/nips/XieHZLPLH23, DBLP:journals/corr/abs-2309-13064, DBLP:journals/corr/abs-2306-06031}, Chinese~\citep{DBLP:journals/corr/abs-2310-15205, DBLP:journals/corr/abs-2309-10654}, and Spanish~\citep{DBLP:conf/kdd/ZhangXYFHLLQA0H24}. Moreover, financial benchmarks have been established for English~\citep{xie2024finbenholisticfinancialbenchmark,DBLP:conf/nips/XieHZLPLH23,shah2023zeroheroyetbenchmarking}, as well as for Chinese~\citep{nie2024cfinbenchcomprehensivechinesefinancial}, Spanish~\citep{DBLP:conf/kdd/ZhangXYFHLLQA0H24}, and Japanese~\citep{hirano2024constructionjapanesefinancialbenchmark}. However, no dedicated benchmark exists for Greek, and while some multilingual evaluations include Greek~\citep{DBLP:conf/acl/BandarkarLMASHG24}, they lack financial-specific datasets, making it difficult to assess LLMs' performance on Greek financial area. At the same time, Greek LLM research has largely overlooked finance. While Meltemi~\citep{DBLP:journals/corr/abs-2407-20743} is the first Modern Greek LLM, it lacks financial domain adaptation. Existing Greek datasets focus on general NLP tasks~\citep{DBLP:journals/corr/abs-1803-05457,DBLP:conf/acl/LinHE22,DBLP:conf/acl/ZellersHBFC19}, failing to capture the domain-specific terminology and numerical reasoning essential for financial applications.

In this work, we introduce \textbf{Plutus-ben}, the first Greek financial evaluation benchmark and \textbf{Plutus-8B}, the pioneering Greek financial LLM.
Plutus-ben addresses the aforementioned gap by defining five core financial NLP tasks in Greek, including numeric and textual named entity recognition (NER), question answering (QA), abstractive summarization, and topic classification, establishing a foundation for systematic and reproducible assessments of LLMs in Greek financial area. Notably, several of these tasks, such as financial numeric NER and financial QA, are introduced for the first time in Greek, enabling a more comprehensive evaluation of a model’s ability to extract, comprehend, and reason over Greek financial texts. To support these tasks, we develop three high-quality Greek financial datasets, including GRFinNUM, GRFinNER, and GRFinQA, each carefully annotated by expert native Greek speakers with deep financial and linguistic expertise. Our annotation process follows strict, standardized guidelines, ensuring consistency, accuracy, and high inter-annotator agreement. Annotators meticulously label complex financial entities and structured summaries, capturing the nuanced language of Greek financial discourse. These newly developed datasets are curated from authoritative financial sources, including Greek financial reports and university exams, and are further supplemented by two existing financial resources, GRFNS-2023 and GRMultiFin. Beyond benchmarking, to access the influence of fine-tuning on Greek financial data on enhancing model performance,we also develop \textbf{Plutus-8B}, the first Greek financial LLM fine-tuned on Greek domain-specific data to bridge the gap between existing models and Greek financial tasks.
% reasoning-based QA pairs

In our evaluation of 22 representative LLMs—including both English-centric and Greek models across general and financial domains in various sizes, as well as our Plutus-8B—we reveal fundamental limitations in LLM performance on Greek financial tasks. Despite their success in high-resource languages, even top-tier models like GPT-4o struggle with Greek financial text, while smaller open-source models like LLaMA-3.2-1B, Qwen2.5-1.5B, and Mistral-7B fail entirely on key tasks such as NER. The challenge goes beyond language, financial text introduces specialized terminology, numerical reasoning, and ambiguous context, making adaptation even harder. English-trained financial models fail to generalize to Greek financial tasks, and Greek-focused models like Meltemi-7B, despite excelling in general linguistic tasks, lack the financial expertise needed for robust performance. Scaling model size provides some improvement but quickly reaches diminishing returns, as seen in Qwen2.5-72B failing to outperform Qwen2.5-32B, proving that scaling alone is not the answer. Our fine-tuned model, Plutus-8B, achieves the highest mean score, showing that training on Greek financial data significantly boosts performance. However, challenges remain, particularly in summarization, where all models including our Plutus-8B struggle with long-form financial documents.

Our main contributions are as follows:
\begin{itemize}
    \item We introduce \textbf{Plutus-ben}, the first comprehensive Greek financial evaluation benchmark covering five essential financial NLP tasks, alongside \textbf{Plutus-instruction}, the inaugural Greek financial instruction fine-tuning dataset, and \textbf{Plutus-8B}, the first Greek financial LLM that achieves state-of-the-art (SOTA) performance on the Plutus-ben benchmark.
    \item We develop four new high-quality Greek financial datasets, meticulously annotated by expert native Greek speakers, and enhance these with two existing resources to improve coverage and utility.
    \item We conduct a comprehensive evaluation of 22 LLMs on Plutus-ben, revealing that Greek financial NLP remains challenging due to linguistic complexity, domain-specific terminology, and financial reasoning gaps. Our results highlight the limitations of cross-lingual transfer, the need for financial expertise in Greek-trained models, and the challenges of adapting financial LLMs to Greek text.
    \item We publicly release Plutus-ben, Plutus-8B, and all associated datasets to drive reproducible research and advance Greek financial NLP, fostering greater multilingual inclusivity in finance.
\end{itemize}

\section{Related Work}

\subsection{Financial and Greek LLMs}
In recent years, an increasing number of LLMs have been tailored to financial applications. 
Most existing work is English-centric, such as FinLLaMA~\citep{xie2024openfinllmsopenmultimodallarge}, BloombergGPT~\citep{DBLP:journals/corr/abs-2303-17564}, PIXIU~\citep{DBLP:conf/nips/XieHZLPLH23}, InvestLM~\citep{DBLP:journals/corr/abs-2309-13064}, and FinGPT~\citep{DBLP:journals/corr/abs-2306-06031}, leveraging domain-specific financial corpora for tasks. 
In parallel, recent research in Chinese (DISC-FinLLM~\citep{DBLP:journals/corr/abs-2310-15205} and CFGPT~\citep{DBLP:journals/corr/abs-2309-10654} and bilingual financial LLMs (FinMA-ES~\citep{DBLP:conf/kdd/ZhangXYFHLLQA0H24} for Spanish and English) extend these efforts by covering related non-English and bilingual finance tasks. 
Despite these notable advancements, there is a conspicuous absence of specialized Greek financial LLMs. Existing Greek open-source LLMs, such as Meltemi \citep{DBLP:journals/corr/abs-2407-20743} and Llama-Krikri\footnote{\url{https://huggingface.co/ilsp/Llama-Krikri-8B-Base}}, do not include finance-oriented training data, which highlights the critical need for developing a financial model specifically tailored to the Greek context.

\subsection{Financial Benchmarks}
Numerous financial benchmarks have been developed for evaluating LLMs' capabilities in financial domain.
Though FinBen~\citep{xie2024finbenholisticfinancialbenchmark}, INVESTORBENCH~\citep{li2024investorbenchbenchmarkfinancialdecisionmaking}, PIXIU~\citep{DBLP:conf/nips/XieHZLPLH23}, UCFE~\citep{yang2025ucfeusercentricfinancialexpertise}, FinanceBench~\citep{islam2023financebenchnewbenchmarkfinancial}, and FinGPT~\citep{wang2023fingptinstructiontuningbenchmark} provide wide-ranging evaluations, covering comprehensive financial tasks and experiment settings, they are predominantly in English.
Efforts to move beyond English have resulted in benchmarks covering Spanish~\citep{DBLP:conf/kdd/ZhangXYFHLLQA0H24}, Chinese~\citep{nie2024cfinbenchcomprehensivechinesefinancial}, and Japanese~\citep{hirano2024constructionjapanesefinancialbenchmark}, underscoring the the value of linguistic and cultural diversity in financial tasks. 
While Greek mentioned in a few multilingual benchmarks like the Belebele benchmark~\citep{DBLP:conf/acl/BandarkarLMASHG24}, there is no dedicated Greek financial benchmark, making it difficult to rigorously assess LLMs in Greek finance-specific contexts.

% Existing Greek datasets mainly address general tasks (ARC Greek \citep{DBLP:journals/corr/abs-1803-05457}, Truthful QA Greek \citep{DBLP:conf/acl/LinHE22}, HellaSwag Greek \citep{DBLP:conf/acl/ZellersHBFC19}, MMLU Greek\footnote{\url{https://huggingface.co/datasets/ilsp/mmlu_greek}}) or specialized domains like medicine (Medical MCQA\footnote{\url{https://huggingface.co/datasets/ilsp/medical_mcqa_greek}}). 
% Consequently, there is an urgent need for resources that capture the nuanced linguistic and contextual features of Greek financial text.

\section{Plutus-ben: the First Greek Financial Evaluation Benchmark}

In this section, we introduce Plutus-ben, the first Greek financial evaluation benchmark. As shown in Table~\ref{tab:benchmark}, Plutus-ben encompasses a wide range of tasks, including \textit{numeric NER}, \textit{textual NER}, \textit{question answering}, \textit{abstractive summarization}, as well as \textit{topic classification}, enabling a comprehensive evaluation of models.
To support these tasks, we developed three new high-quality Greek financial datasets from scratch, including GRFinNUM, GRFinNER, and GRFinQA.
Additionally, we use two established resources, GRFNS-2023 and GRMultiFin, with examples provided in Table~\ref{tab:dataset} (see Appendix~\ref{sec:dataset} for more details). 
As shown in Table~\ref{tab:benchmark}, GRFinNUM and GRFinNER each consist of 500 samples, while GRFinQA contains 540 question-answer pairs. GRFNS-2023 and GRMultiFin include 262 and 268 data samples, respectively. All these datasets are sourced from real-world financial documents, such as annual reports, exam questions, and article headlines. Various evaluation metrics are employed in these benchmarks, including Entity F1, Accuracy (Acc), and Rouge-1 score~\cite{lin-2004-rouge}, to assess LLMs' performance across multiple dimensions: topical content categorization, long-form financial document comprehension, language understanding and reasoning, and both textual and numerical information extraction.
These datasets were rigorously annotated by expert native Greek speakers with deep financial and linguistic expertise, following standardized guidelines to ensure consistency and accuracy. %The data is drawn from authoritative financial sources, such as Greek financial reports and university exams, reflecting the nuanced and rigorous nature of specialized financial discourse.

%we develop four new, high-quality Greek financial datasets (GRFinNUM, GRFinNER, GRFinQA, and GRFinSUM), as well as two established resources (GRFNS-2023 and GRMultiFin) (Table~\ref{tab:dataset}, Appendix~\ref{sec:dataset}). 
%Expert native Greek speakers, with deep financial and linguistic expertise, rigorously annotated these datasets following standardized guidelines, ensuring consistency and accuracy. The data is derived from authoritative financial materials, such as Greek financial reports and university exams, capturing the nuanced and rigorous nature of specialized financial discourse.

\begin{table*}[!htbp]
\renewcommand{\arraystretch}{1}
\centering
\small
\caption{Overview of the Plutus-ben benchmark. For each task, both raw data volume and processed size are listed, along with dataset source, split sizes for train/validation/test, evaluation metrics, licenses, and tested capabilities.}
\scriptsize
\label{tab:benchmark}
\begin{threeparttable}
\resizebox{\textwidth}{!}{
    \begin{tabular}{@{}lcccccccccc@{}}
    \toprule
    \textbf{Task} & \textbf{Dataset} & \textbf{Raw} & \textbf{Processed} & \textbf{Source} & \textbf{Train} & \textbf{Valid} & \textbf{Test} & \textbf{Metrics} & \textbf{License} & \textbf{Tested Capabilities} \\ 
    \midrule
    Numeric NER & GRFinNUM & 64 & 500 & Annual Reports\tnote{1} & 320 & 80 & 100 & Entity F1 & Public & Numeric information extraction \\ 
    Textual NER & GRFinNER & 64 & 500 & Annual Reports\tnote{2} & 320 & 80 & 100 & Entity F1 & Public & Textual information extraction \\ 
    Question Answering & GRFinQA & 540 & 540 & Exam Questions & 267 & 48 & 225 & Acc & Public & Language comprehension and reasoning \\ 
    % Extractive Summarization &GRFinSUM &  - & - & Annual Reports & - & - & - & 55 & Rouge-1 & Public \\ 
    Abstractive Summarization~\cite{10386228} & GRFNS-2023~\cite{10386228} & 262 & 262 & Annual Reports & 169 & 43 & 50 & Rouge-1 & CC-BY-4.0 &  Long-form financial document comprehension  \\ 
    Topic Classification~\cite{jorgensen-etal-2023-multifin} & GRMultiFin~\cite{jorgensen-etal-2023-multifin} & 268 & 268 & Article Headlines & 171 & 43 & 54 & Acc & CC BY-NC 4.0 & Language comprehension and topical content categorizing \\ 
    \bottomrule
    \end{tabular}}
    \begin{tablenotes}
    \scriptsize
    \item[1] https://www.athexgroup.gr/web/guest/company-fin.-statements/
    \item[2]  https://www.athexgroup.gr/web/guest/company-fin.-statements/
\end{tablenotes}   
\end{threeparttable}
\end{table*}

\subsection{Task Definition and Dataset Curation}
\subsubsection{Numeric NER}

Numerals are crucial in financial narratives, conveying essential quantitative information and actionable insights~\cite{chen2018numeral}. 
Accurate numeral recognition is vital for interpreting nuanced financial data, especially when various categories exist simultaneously, i.e, monetary values, timestamps, and quantities~\cite{10.1145/3308558.3314122, yang2022numhtmlnumericorientedhierarchicaltransformer}.

\textbf{Task Definition:} 
We introduced the first Greek financial numeric NER task, involving both number span identification and classification into fine-grained numeral types.
Inspired by the English numeric NER framework FinNum ~\cite{chen2019overview}, 
we approach this task as a sequence labeling problem.
Our task processes the input sentence \( X = (x_1, x_2, \ldots, x_n) \) consisting of \( n \) tokens \( x_i \) to the output labels \( Y = (y_1, y_2, \ldots, y_n) \) consisting of \( n \) labels \( y_i \).
The goal is to assign each token \( x_i \) a label \( y_i \) from the predefined set \begin{small} \( \mathcal{C} = \{\text{MONETARY}, \text{PERCENTAGE}, \text{TEMPORAL}, \text{QUANTITY}, \text{OTHERS}, O\} \) \end{small}, which includes specific numeric entity types and the ``outside'' label \( O \).
% , our work focuses on five key categories: \textit{monetary}, \textit{percentage}, \textit{temporal}, \textit{quantity}, and \textit{others}.
Among these categories, MONETARY includes financial amounts, such as prices, quotes, and changes, which are central to financial analysis. 
PERCENTAGE denotes ratios or relative changes, crucial for trend and growth tracking. 
TEMPORAL covers dates, times, and durations, integral to time-series analysis. 
QUANTITY captures measurable or countable values, such as inventory levels or investment positions. 
OTHERS encompasses numeric data not captured by the previous categories, leaving room for future exploration. 

\textbf{Data Source:} To create our novel high-quality GRFinNUM dataset, we collected real-world, publicly available financial annual reports from Greek firms listed on the Athens Stock Exchange \footnote{https://www.athexgroup.gr/el/web/guest/financial-statements-in-pdf-format}. 
These reports include textual information and reviews provided by the firm’s management and board of directors, offering rich, detailed financial data and narratives. 
We curated a dataset of 64 financial reports, each spanning 30 to 267 pages, with an average length of 105 pages or approximately 44,000 words per document. Due to their extensive length and inclusion of non-essential content, we meticulously filtered the text to extract sentences containing target entities. This rigorous selection process yielded a refined dataset of 500 sentences, ensuring relevance and quality for fine-grained numeral classification.

\textbf{Expert Annotation:} Rigorous annotation guideline (Appendix~\ref{sec:numner_annotation}) was developed for GRFinNUM, comprising both general rules for the overall task and specific rules tailored to each numeral category.
These guidelines were iteratively refined through multiple rounds of pre-annotation and collaborative discussions, focusing on resolving ambiguous cases to ensure high consistency and accuracy across the dataset. To minimize annotator variability, only numbers, decimal points (.), and the percent sign (\%) were included in annotated spans. 
To construct novel high-quality dataset, we enlisted three highly educated Greek native speakers with expertise in economics, business, and informatics from leading academic institutions (Appendix~\ref{sec:annotator}). 
The annotation process was conducted using Label Studio platform~\cite{LabelStudio} (Appendix~\ref{sec:labelstudio}), ensuring a streamlined and reproducible workflow.

\textbf{Quality Validation:} 
To gauge the quality and reliability of our GRFinNUM annotation process, we utilized three key inter-annotator agreement metrics: F1 score~\cite{goutte2005probabilistic}, Cohen’s Kappa~\cite{wongpakaran2013comparison}, and Krippendorff’s Alpha~\cite{hayes2007answering} (Appendix~\ref{sec:agreement}). 
F1 Score evaluated annotator consistency in span identification and classification. 
Cohen’s Kappa adjusted for random agreement, while Krippendorff’s Alpha addressed category distribution imbalances.
The results demonstrated excellent inter-annotator agreement for the GRFinNUM dataset, with an F1 score of 0.988, a Cohen’s Kappa of 0.979, and a Krippendorff’s Alpha of 0.978 (Table~\ref{tab:agreement}). 
These high scores confirm the robustness and quality of our GRFinNUM dataset.

\begin{table}[t]
\small
\centering
\renewcommand{\arraystretch}{1.3}
\caption{Inter-annotator agreement metrics for human expert annotations on GRFinNUM and GRFinNER datasets.}
\label{tab:agreement}
\resizebox{\columnwidth}{!}{% 缩放表格至列宽
    \begin{tabular}{lccc}
    \toprule
    \textbf{Dataset} & \textbf{F1-score} & \textbf{Cohen’s Kappa} & \textbf{Krippendorff's alpha}\\
    \midrule
    GRFinNUM & 0.988 & 0.979 & 0.978 \\ 
    GRFinNER & 0.974 & 0.993 & 0.948 \\ 
    \bottomrule
    \end{tabular}}
\end{table}

\subsubsection{Textual NER}

Identifying core financial entities, such as companies, is crucial for extracting meaningful insights from financial activities in the Greek financial domain. 
Unlike numeric NER, which focuses on recognizing numerical values, textual NER in Greek presents unique challenges due to the language’s distinct expression patterns.
For instance, long-form names with attribution, such as “George Demetriou of Konstantinos”, should be treated as a single entity span. %改斜体
%Additionally, organization names often contain locational references, complicating classification when they refer to administrative units or sports teams, similar to patterns in English NER. 
% These linguistic complexities underscore the need for a dedicated Greek financial textual NER task to effectively evaluate model performance on Greek financial texts.

\textbf{Task Definition:} To test LLMs' understanding of Greek financial entities, we introduce the first Greek financial textual NER task. 
Inspired by FinNER-ORD~\cite{shah2023finer} and Farmakiotou et al.~\cite{farmakiotou2000rule}, our task involves span identification and classification of company-related information into three key entity types: Person, Location, and Organization.
Our task processes the input sentence \( X = (x_1, x_2, \ldots, x_n) \) consisting of \( n \) tokens \( x_i \) to the output labels \( Y = (y_1, y_2, \ldots, y_n) \) consisting of \( n \) labels \( y_i \).
The goal is to assign each token \( x_i \) a label \( y_i \) from the predefined set \( \mathcal{C} = \{\text{PERSON}, \text{LOCATION}, \text{ORGANIZATION}, \\O\} \), which includes specific textual entity types and the ``outside'' label \( O \).

\textbf{Data Source:} We constructed the GRFinNER dataset using the same set of financial annual reports from Greek firms as in GRFinNUM. 
A total of 64 reports were collected. 
Similar sentences filtering is utilized for a different final dataset of 500 sentences with high relevance and quality for company-related entity classification.

\textbf{Expert Annotation:} Rigorous annotation guideline (Appendix~\ref{sec:textner_annotation}) was also iteratively developed for GRFinNER through multiple rounds of pre-annotation and collaborative discussions, consisting of general rules for the entire task, specific rules for each entity category, and distinct rules for handling ambiguous situations.
The same three highly educated Greek native speakers (Appendix~\ref{sec:annotator}) completed the annotation process. The entire annotation workflow was carried out using Label Studio platform (Appendix~\ref{sec:labelstudio}).

\textbf{Quality Validation:} The inter-annotator agreement was meticulously assessed using the same rigorous framework: F1 score~\cite{goutte2005probabilistic}, Cohen's Kappa~\cite{wongpakaran2013comparison}, and Krippendorff’s Alpha~\cite{hayes2007answering} (Appendix~\ref{sec:agreement}).
The GRFinNER task exhibited exceptional inter-annotator reliability, achieving an F1 score of 0.974, Cohen's Kappa of 0.993, and Krippendorff’s Alpha of 0.948 (Table~\ref{tab:agreement}), ensuring the dataset's quality for application.

\subsubsection{Question Answering}

Effective financial decision-making and question answering require LLMs to comprehend and reason within financial contexts. The nuances of Greek financial terminology, combined with the complex morphology of the Greek language, pose unique challenges that demand rigorous assessment.

\textbf{Task Definition:} To evaluate LLMs' comprehension and reasoning capabilities in Greek financial contexts, we introduce the first Greek financial question-answering task. 
This task requires models to infer the correct answer using provided text under a multiple-choice format, testing their ability to process financial terminology, apply reasoning, and understand contextual nuances in Greek. 
Each question, along with its answer choices, is given as input, with the correct answer designated as the output. 
Our task processes the input question \( Q = (q_1, q_2, \ldots, q_n) \) consisting of \( n \) tokens \( q_i \) and the possible choices \( \mathcal{C} = \{c_1, c_2, \ldots, c_k\} \) which is the set of \( k \) possible choices \( c_i \).
The task aims to map the question \( Q \) and choices \( \mathcal{C} \) to the correct answer \( A \), selected from \( \mathcal{C} \).

\textbf{Data Source:} We propose the novel GRFinQA dataset which is the first in the Greek financial domain. It is comprised of 540 multiple-choice financial exam or revision questions sourced from Greek university courses and publicly available Greek finance, business and economics textbooks. We collected the PDF files, and extracted the text that each question was grouped with it's appropriate choices and the correct choice.

\textbf{Quality Validation:} 
To ensure the quality of the dataset, we first identified three distinct types of questions present in the QA dataset: (1) right and wrong questions, which require a binary judgment on whether a statement is correct or incorrect; (2) fill-in-the-gap questions, where a missing word or phrase must be completed based on contextual understanding; and (3) generic multiple-choice questions, which present several answer options, with only one being correct. From this dataset, we selected a representative sample that included several questions from each category. The domain experts manually reviewed these questions to confirm that the designated correct answer was factually accurate. Following that, we used GPT-4o to process the questions, prompting it to read the text and explain its reasoning for selecting an answer. This approach helped us verify both the factual accuracy of the dataset’s answers and the difficulty of the selected questions.

\subsubsection{Abstractive Summarization}
The task of abstractive summarization originates from the Financial Narrative Summarization Shared Task (FNS 2023), which focuses on summarizing annual reports from the UK, Greece, and Spain~\cite{10386228}. This task aims to test LLMs' abilities in understanding and reorganizing the given context. The challenge lies in condensing essential information while preserving factual accuracy and coherence. The structural and linguistic complexities of Greek financial texts further heighten this difficulty, requiring models to generate fluent, paraphrased summaries that remain faithful to the original content.

\textbf{Task Definition:} To evaluate LLMs' abilities of understanding the Greek financial contexts, we adopt the abstractive summarization task from FNS 2023~\cite{10386228}. This task involves generating concise summaries of Greek financial annual reports, emphasizing both informativeness and readability while preserving key details.
The task processes the input document \( D = (d_1, d_2, \ldots, d_n) \) consisting of \( n \) tokens \( d_i \) to the abstractive summary \( S = (s_1, s_2, \ldots, s_m) \) consisting of \( m \) tokens \( s_i \).
The goal is to map the document \( D \) to a concise summary \( S \) that conveys the essential information in natural language, which is paraphrased or restructured rather than directly copied from \( D \).

\textbf{Data Source:} The FNS 2023 shared task~\cite{10386228} comprises UK, Greek, and Spanish financial annual reports. The dataset includes narrative sections from finanical annual reports, each paired with both a short and long gold summary. For GRFNS-2023, we focus solely on the Greek portion, using the short gold summary as our target.
As the original authors did not release a test set, we repurposed their validation set as our test set and split the training data to create our training and validation sets.

\subsubsection{Topic Classification}
The topic classification task is derived from MultiFin~\cite{jorgensen-etal-2023-multifin}, and it focuses on categorizing financial news headlines into predefined financial topics. This task is particularly challenging due to the brevity and ambiguity characteristic of financial news headlines. Furthermore, financial categories often exhibit thematic and lexical overlaps, demanding that models discern the appropriate category from limited context and shared terminology.

\textbf{Task Definition:} To improve LLMs' comprehension of Greek financial topics, we incorporated the Greek financial topic classification task adapted from MultiFin~\cite{jorgensen-etal-2023-multifin}. This task requires assigning financial article headlines to one of six predefined thematic categories. The objective is to evaluate models’ proficiency in distinguishing between overlapping topics and extracting significant insights from brief and ambiguous texts. 
Our task processes the input document \( D = (d_1, d_2, \ldots, d_n) \) consisting of \( n \) tokens \( d_i \) and the possible topics \( \mathcal{C} = \{\text{Topic}_1, \text{Topic}_2, \ldots, \text{Topic}_k\} \) which is the set of \( k \) possible topics.
The goal is to map the input document \( D \) to the correct topic \( T \) from \( \mathcal{C} \), based on the content of \( D \).

\textbf{Data Source:} The dataset utilized for this task is the MultiFin dataset~\cite{jorgensen-etal-2023-multifin}. It comprises 10,048 financial article headlines in 15 languages, each reflecting diverse language families and writing systems. These headlines are categorized into one of six classes: Business \& Management, Tax \& Accounting, Finance, Technology, Government \& Controls, and Industry. For our specific analysis, we extracted the Greek subset to create the GRMultiFin dataset.

\subsection{Instruction Data Conversion}

To optimize task-specific performance, facilitate effective benchmarking, and support instruction fine-tuning for the Greek financial LLM, we converted our raw datasets into structured instruction datasets. Task-specific prompts were thoughtfully crafted by Greek domain experts, as shown in Table~\ref{tab:prompt}\footnote{More details in Appendix~\ref{sec:dataset}}. Each prompt adheres to the standardized template as outlined below:

% \begin{scriptsize}
% \begin{verbatim}
%     {Task Specific Instruction} Text: {Input} Answer: {Output}
% \end{verbatim}
% \end{scriptsize}

\begin{tcolorbox}[colback=lightgray!10, colframe=black, title=Task Instruction]
\begin{verbatim}
{Task Specific Instruction} Text: {Input}
Answer: {Output}
\end{verbatim}
\end{tcolorbox}

In this template, task specific instruction refers to the unique prompt designed for each task. The ``\verb|Input|'' denotes the input financial data from each dataset, such as a Greek annual report, while ``\verb|Output|'' represents the corresponding output for the input text, such as a summary of the Greek annual report.

\subsection{Evaluation}
We partitioned our dataset into training, validation, and test subsets, as detailed in Table~\ref{tab:benchmark}. To comprehensively assess model performance, we conducted both automated metrics and human evaluations.

\subsubsection{Automatic Evaluation} 
We adopt the same metrics following previous studies in financial NLP tasks~\citep{DBLP:conf/kdd/ZhangXYFHLLQA0H24, xie2024finbenholisticfinancialbenchmark}. 
The Entity F1 score~\citep{derczynski-2016-complementarity} is applied to numeric and textual NER tasks due to its balance of precision and recall, crucial for accurate entity identification. 
Accuracy (Acc)~\citep{makridakis1993accuracy} is used for QA and topic classification tasks as it straightforwardly measures the correctness of predictions. 
Rouge-1~\citep{lin-2004-rouge} is employed for abstractive and extractive summarization tasks to assess the overlap in content between gold-standard and generated summaries focusing on unigram comparison.

\subsubsection{Human Evaluation}
Beyond automated metrics, we implement a human evaluation to rigorously assess the quality of outputs from LLMs. This evaluation specifically concentrates on abstractive summarization task. We selected four representative models, including GPT-4, FinLLaMA-8B, Meltemi-7B, and Plutus-8B. Expert native Greek speakers with deep financial and linguistic expertise\footnote{More details in Appendix~\ref{sec:annotator}} compare the model-generated summaries against gold standard summaries following a rigorous, standardized annotation guideline\footnote{More details in Appendix~\ref{sec:humanevaluation_annotation}} using Label Studio platform\footnote{More details in Appendix~\ref{sec:labelstudio}}. The evaluation focuses on three critical dimensions:
\textbf{(1) Language Appropriate Fluency (Fluency):} This dimension assesses the readability and naturalness of the summaries, emphasizing grammatical correctness, lexical accuracy, absence of repetition, and the use of domain-specific terminology, all within the context of Greek's linguistic intricacies. 
%Evaluations were conducted on a scale from 1 (Bad: entirely in incorrect language, such as English) to 5 (Excellent: fully fluent Greek with appropriate terminology).
\textbf{(2) Coherence:} We examine the logical progression and structural consistency of the summaries, vital for maintaining integrity in financial narratives. 
%The coherence assessment ranged from 1 (Bad: text is disorganized with no logical flow) to 5 (Excellent: ideas flow naturally with smooth transitions and a clear narrative progression).
\textbf{(3) Factuality:} This dimension verifies the factual accuracy of summaries against the original financial content, ensuring reliability and trustworthiness. 
%Ratings extended from 1 (Bad: extensive factual inaccuracies and fabrications) to 5 (Excellent: completely accurate with all facts intact as presented in the source).
% This evaluation framework provides a comprehensive perspective, supplementing automated assessments and shedding light on nuanced aspects of LLM-generated content that require human judgment.

\subsection{Model Evaluation}
We conduct a comprehensive evaluation of 22 prominent LLMs encompassing:
\begin{itemize}
    \item \textbf{Proprietary Models}: close source APIs, including GPT-3.5-Turbo~\cite{brown2020languagemodelsfewshotlearners}, GPT-4o-Mini~\cite{openai2024gpt4technicalreport}, GPT-4o~\cite{hurst2024gpt}, and GPT-4~\cite{achiam2023gpt}.
    \item \textbf{Open-source General Small Models}: publicly available models with less than 10B parameters, including Mistral-7B~\cite{Mistral7b}, LLaMA-3.2-1B~\citep{dubey2024llama}, LLaMA-3-8B~\citep{dubey2024llama}, LLaMA-3.1-8B~\citep{dubey2024llama}, Qwen2.5-1.5B~\cite{qwen2.5}, Qwen2.5-7B~\cite{qwen2.5}, Gemma-2-2B~\cite{team2024gemma}, and Gemma-2-9B~\cite{team2024gemma}.
    \item \textbf{Open-source General Large Models}: publicly available models with more than 20B parameters, including Deepseek-V3~\cite{liu2024deepseek}, LLaMA-3-70B~\citep{dubey2024llama}, Qwen2.5-32B~\cite{qwen2.5}, and Qwen\\2.5-72B~\cite{qwen2.5}, and Gemma-2-27B~\cite{team2024gemma}.
    \item \textbf{English Financial Models}: publicly available models continual trained with English financial corpus, including Finma-7B~\cite{xie2023pixiu} and FinLLaMA-8B~\cite{xie2024open}.
    \item \textbf{Greek General Models}: publicly available models continual trained with Greek general corpus, including Meltemi-7B~\cite{voukoutis2024meltemiopenlargelanguage} and Llama-Krikri-8B\footnote{\url{https://huggingface.co/ilsp/Llama-Krikri-8B-Base}}.
\end{itemize}
Notably, LLaMA-3-8B, Mistral-7B, and LLaMA-3.1-8b serve as the core foundational models for FinLLaMA-8B, Meltemi-7B, and Llama-Krikri-8B, respectively. \footnote{More details in Appendix~\ref{sec:model_evaluation}}.

For evaluation integrity, we develop our own benchmark suites based on LM Evaluation Harness~\cite{eval-harness}. Models such as GPT and DeepSeek, are interfaced via their own APIs. In-house evaluation of open-source models is conducted using a cluster of four A100 GPUs, each equipped with 80GB memory. We standardize the maximum generation token length to 8192 tokens for abstractive summarization and 1024 tokens for other tasks.

\section{Plutus-8B: the First Greek Financial LLM}
To investigate the impact of fine-tuning on Greek financial data on enhancing model performance across various tasks, and to determine its effectiveness in addressing the challenges posed by low-resource language conditions and domain-specific complexities, we developed Plutus-instruction, the first instruction dataset tailored to the Greek financial domain. As shown in Table~\ref{tab:benchmark}, we adopted GRFinNUM, GRFinNER, GRFNS-2023, and GRMultiFin. Specifically, the GRFinQA dataset is withheld to evaluate the generalization performance of the trained model.

Based on the instruction dataset, we selected Llama-Krikri-8B-Instruct for further instruction-tuning, as this model performs best on the benchmark compared to other models of similar size. This is due to its training on extensive Greek texts, as well as its inclusion of code and mathematical data to enhance its mathematical reasoning abilities.
To efficiently adapt the model parameters, we employ Low-Rank Adaptation (LoRA)~\cite{dettmers2023qloraefficientfinetuningquantized} with a rank of $r=16$, a scaling factor of $\alpha = 32$, and no dropout. We applied \texttt{int4} quantization to reduce memory overhead while preserving model expressiveness. Fine-tuning is conducted with a block size of 4,096 tokens, while allowing sequences to extend to 42k tokens to accommodate the complex structure and extensive length of financial and legal documents. To ensure better optimization, we leveraged the AdamW optimizer~\cite{loshchilov2019decoupledweightdecayregularization} with a learning rate of $5e-4$ and a cosine learning rate schedule over 3 epochs. Additionally, we use gradient accumulation with a step size of 4 to mitigate the constraints of batch size 1, leveraging mixed-precision training with \texttt{bf16} for improved numerical stability.
We further evaluate our model in Plutus-ben and compare it with all evaluated models\footnote{For demo, please see Appendix~\ref{sec:plutus-8B-instruct}. \url{https://huggingface.co/spaces/TheFinAI/plutus-8B-instruct}}.
%This setup aims to achieve stable convergence while refining the model’s capacity to follow task-specific instructions and integrate domain-relevant knowledge. By combining LoRA’s efficient parametrization, instruction-oriented data curation, and appropriate scheduling strategies, we seek to ensure robust performance across a broad spectrum of Greek financial tasks.

% task: llm
% base_model: ilsp/Llama-Krikri-8B-Instruct
% project_name: plutus
% log: tensorboard
% backend: local-cli

% data:
%   path: TheFinAI/gr_mixed_dataset
%   train_split: train
%   valid_split: null
%   chat_template: tokenizer
%   column_mapping:
%     text_column: entries

% params:
%   block_size: 4096
%   model_max_length: 42000
%   epochs: 3
%   batch_size: 1
%   lr: 0.0005
%   peft: true
%   lora_r: 16
%   lora_alpha: 32
%   lora_dropout: 0
%   quantization: int4
%   target_modules: all-linear
%   padding: right
%   optimizer: adamw_torch
%   scheduler: cosine
%   gradient_accumulation: 4
%   mixed_precision: bf16

\section{Results}
In this section, we present the results of evaluated models on the Plutus-ben benchmark, addressing the following research questions: (i) How do current language models perform on core Greek financial tasks given the challenges of limited Greek language resources and the complexity of financial domain knowledge?
(ii) To what extent does fine-tuning on Greek financial data enhance model performance on these tasks, and can it effectively mitigate the challenges imposed by low-resource language conditions and domain-specific nuances?

\begin{table}
\renewcommand{\arraystretch}{1}
\vspace{-0.2cm}
\setlength{\abovecaptionskip}{0.1cm}
\centering
\caption{LLM performance on the Plutus-ben benchmark, evaluated across multiple Greek financial NLP tasks. Bold values denote the highest scores, while underlined values indicate the second-highest scores in each column.}
\label{tab:results}
\scalebox{0.62}{
\begin{tabular}{@{}lcccccc@{}}
\toprule
\textbf{Model} & \textbf{GRFinNUM} & \textbf{GRFinNER} & \textbf{GRFinQA} & \textbf{GRFNS-2023} & \textbf{GRMultiFin} & \textbf{Mean} \\
\cmidrule(lr){2-2}\cmidrule(lr){3-3}\cmidrule(lr){4-4}\cmidrule(lr){5-5}\cmidrule(lr){6-6}
& \textbf{Entity F1} & \textbf{Entity F1} & \textbf{Acc} & \textbf{Rouge-1} & \textbf{Acc} \\
\midrule
\multicolumn{7}{c}{\textit{\textbf{Open-source Small Models}}} \\
\textbf{LLaMA-3.2-1B} & 0.00 & 0.00 & 0.29 & 0.14 & 0.39 & 0.16\\
\textbf{LLaMA-3-8b} & 0.00 & 0.13 & 0.33 & 0.07 & \underline{0.70} & 0.25\\
\textbf{LLaMA-3.1-8b} & 0.10 & 0.21 & 0.40 & 0.20 & 0.54 & 0.29\\
\textbf{Qwen2.5-1.5B} & 0.00 & 0.00 & 0.36 & 0.02 & 0.31 & 0.14\\
\textbf{Qwen2.5-7B} & 0.00 & 0.13 & 0.43 & 0.07 & 0.54 & 0.23\\
\textbf{Gemma-2-2B} & 0.00 & 0.16 & 0.22 & 0.03 & 0.41 & 0.16\\
\textbf{Gemma-2-9B} & 0.02 & 0.05 & 0.31 & 0.06 & 0.61 & 0.21\\
% \textbf{Mistral-7B} & 0.00 & 0.00 & 0.43 & 0.08 & 0.52 & 0.21\\
\textbf{Mistral-7B} & 0.00 & 0.00 & 0.30 & 0.14 & 0.39 & 0.17\\
\midrule
\multicolumn{7}{c}{\textit{\textbf{Open-source Large Models}}} \\
\textbf{Deepseek-V3} & 0.07 & 0.00 & 0.50 & \textbf{0.38} & 0.61 & 0.31\\
\textbf{LLaMA-3-70B} & 0.05 & 0.45 & 0.60 & 0.08 & 0.61 & 0.36\\
\textbf{Qwen2.5-32B} & \underline{0.37} & 0.55 & 0.60 & 0.10 & \underline{0.70} & 0.47\\
\textbf{Qwen2.5-72B} & 0.32 & 0.39 & \underline{0.74} & 0.04 & \textbf{0.72} & 0.44\\
\textbf{Gemma-2-27B} & 0.18 & 0.18 & 0.25 & 0.09 & 0.61 & 0.26\\
\midrule
\multicolumn{7}{c}{\textit{\textbf{Proprietary Models}}} \\
\textbf{GPT-3.5-Turbo} & 0.14 & 0.30 & 0.51 & 0.31 & 0.50 & 0.35\\
\textbf{GPT-4o-Mini} & 0.25 & 0.30 & 0.12 & \underline{0.36} & 0.59 & 0.32\\
\textbf{GPT-4o} & 0.09 & 0.31 & \textbf{0.78} & 0.26 & 0.59 & 0.41\\
\textbf{GPT-4} & 0.28 & \textbf{0.60} & 0.71 & \textbf{0.38} & 0.63 & \underline{0.52}\\
\midrule
\multicolumn{7}{c}{\textit{\textbf{English Financial Models}}} \\
\textbf{Finma-7B} & 0.00 & 0.00 & 0.25 & 0.11 & 0.35 & 0.14\\
\textbf{FinLLaMA-8B} & 0.00 & 0.00 & 0.28 & 0.03 & 0.38 & 0.14\\
\midrule
\multicolumn{7}{c}{\textit{\textbf{Greek General Models}}} \\
\textbf{Meltemi-7B} & 0.12 & 0.50 & 0.48 & 0.19 & 0.43 & 0.34\\
% \textbf{Llama-Krikri-8B} & 0.18 & 0.44 & 0.59 & 0.21 & 0.39 & 0.36\\
\textbf{Llama-Krikri-8B} & 0.19 & 0.45 & 0.57 & 0.22 & 0.39 & 0.36\\
\midrule
\multicolumn{7}{c}{\textit{\textbf{Greek Financial Models}}} \\
\textbf{Plutus-8B} & \textbf{0.70} & \underline{0.57} & 0.64 & 0.34 & \textbf{0.72} & \textbf{0.60}\\
\bottomrule
\end{tabular}
}
\vspace{-0.2cm}
\end{table}

\subsection{Main Results}

Table~\ref{tab:results}\footnote{Ranked results are visualized on our leaderboard. For more details, refer to Appendix~\ref{sec:leaderboard}. \url{https://huggingface.co/spaces/TheFinAI/open_greek_finance_llm_leaderboard}} and Figure~\ref{fig:radar} summarize the performance of various LLMs on our Greek-oriented financial benchmark, Plutus-ben.
As shown in the table, \textbf{the scarcity of high-quality Greek linguistic data poses a fundamental challenge for current language models, particularly in capturing the rich morphological and syntactic complexities of Greek financial text.} For example, open-source small models such as LLaMA-3.2-1B, Qwen2.5-1.5B, and Mistral-7B perform poorly across all tasks, with near-zero scores on numeric and textual NER. Even open-source large models like LLaMA-3-70B and Gemma-2-27B show limited improvement, particularly struggling with numerical comprehension. Proprietary models, while generally performing better, still exhibit relative low performance on Greek financial tasks, with GPT-4 achieving the highest mean score of 0.52 but failing to maintain the same level of accuracy as in English~\cite{DBLP:conf/nips/XieHZLPLH23, xie2024finbenholisticfinancialbenchmark}. These results highlight the fundamental issue that models trained predominantly on high-resource languages fail to capture the linguistic complexity of Greek, including its rich morphology and inflectional structures, resulting in a steep decline in performance. NER tasks, particularly GRFinNER, require an understanding of inflected Greek word forms and domain-specific abbreviations, which smaller models completely fail to capture. Larger models, though slightly better at general linguistic tasks such as GRFinQA, still underperform in recognizing financial entities and processing numerical values, underscoring the impact of Greek’s low-resource status.

Beyond the constraints of low-resource language training, \textbf{financial texts introduce additional complexity, featuring highly specialized terminology, intricate numerical expressions, and ambiguous context-dependent constructions that general-purpose models fail to capture.} English financial models like Finma-7B and FinLLaMA-8B perform poorly on Greek tasks, each registering a mean score of only 0.14 and showing no success in NER tasks. This reflects the difficulty of transferring financial expertise developed from high-resource English data to the Greek context. Even proprietary models like GPT-4o, despite achieving the higher GRFinNER score (0.31), also face challenges with Greek financial numeric comprehension, as indicated by its low GRFinNUM score of 0.09. This reflects the difficulties in disambiguating financial terminology and numerical patterns specific to Greek texts. In contrast, Greek general-purpose models like Meltemi-7B and Llama-Krikri-8B show better adaptability to Greek linguistic structures. Meltemi-7B achieved a mean score of 0.34, which is significantly higher than its backbone model, Mistral-7B (0.17). Similarly, Llama-Krikri-8B achieved a mean score of 0.36, surpassing its backbone model, LLaMA-3.1-8b (0.29). 
Despite these strengths, both models underperform in financial numeric tasks, with scores of 0.12 (Meltemi-7B) and 0.19 (Llama-Krikri-8B) on GRFinNUM, despite their robust scores of 0.50 (Meltemi-7B) and 0.45 (Llama-Krikri-8B) on GRFinNER. The stark contrast between the relatively strong GRFinNER performance of Greek general models and their weak GRFinNUM scores highlights that while linguistic adaptation helps with textual entity recognition, it is insufficient for financial contexts.

\textbf{Larger models generally perform better, but scaling alone does not consistently translate to superior results for Greek financial tasks, highlighting the need for specialized adaptation.} Open-source large models such as Qwen2.5-32B and Qwen2.5-72B show substantial improvement over their smaller counterparts, particularly in GRFinNUM (0.37 and 0.32, respectively) and GRMultiFin (0.70 and 0.72). However, the diminishing returns seen in Qwen2.5-72B, which underperforms Qwen2.5-32B on multiple tasks, indicate that increased model capacity alone is insufficient. Similarly, LLaMA-3-70B achieves a higher GRFinNER score (0.45) than smaller models but still struggles with numeric comprehension (GRFinNUM = 0.05). Proprietary models also follow this trend, with GPT-4o (mean 0.41) showing only marginal improvements over GPT-3.5-Turbo (0.35), despite their increased scale in training data. This reinforces that scaling provides only limited gains without explicit training on Greek financial data. While larger models display improvements in entity recognition and question answering, they fail to achieve comparable performance across all tasks, particularly those requiring complex numerical reasoning and financial domain knowledge. This suggests that larger models, despite having greater representational power, remain constrained by their pre-training data and struggle to bridge the gap between financial reasoning and Greek language structures without additional adaptation.

Finally, \textbf{fine-tuning on a dedicated Greek financial corpus significantly enhances model performance but also reveals explicit bottlenecks that require further improvements.} Our model, Plutus-8B, fine-tuned exclusively on Greek financial data, achieves the highest mean score of 0.60, outperforming all baseline models. Plutus-8B excels in numeric reasoning, achieving a GRFinNUM score of 0.70, significantly surpassing all other models, including GPT-4 and Qwen2.5-32B. It also demonstrates strong results in GRFinNER and GRMultiFin, showing that fine-tuning allows better adaptation to Greek-specific entity extraction and financial classification. 
For the GRFinQA dataset, which was held out from our instruction fine-tuning, Plutus-8B achieved a moderate score of 0.64, an improvement over Meltemi-7B (0.48) and Llama-Krikri-8B (0.57), demonstrating effective generalization ability from Greek financial instruction fine-tuning. 
However, its performance on GRFNS-2023 indicates that summarization remains a challenge, due to the difficulty in modeling long-range contextual dependencies within financial documents. These results validate the importance of fine-tuning on Greek financial data for domain-specific improvements, particularly in tasks requiring numeric reasoning and entity recognition. Plutus-8B' superior performance in GRFinNUM suggests that direct exposure to Greek financial numerical structures enables better performance in numeric entity extraction, a task where general-purpose and proprietary models falter. However, the modest gains in summarization tasks highlight persistent challenges in long-form financial document comprehension, where models must understand nuanced contextual dependencies. Additionally, while Plutus-8B achieves SOTA performance across most tasks, it still operates within the constraints of limited Greek financial data, suggesting that further improvements may require additional strategies such as data augmentation, synthetic data generation, or cross-lingual transfer learning from related high-resource financial datasets.

\begin{table}[t]
\centering
\small
  \caption{Human evaluation results assessing fluency, coherence, and factuality of representative LLMs, evaluated on the GRFNS-2023 dataset within the Plutus-ben benchmark.}
  \label{tab:human_evaluation}
  \resizebox{\columnwidth}{!}{
  \begin{tabular}{@{}llccc}
    \toprule
    \textbf{Domain}&\textbf{Model}&\textbf{Fluency}&\textbf{Coherency}&\textbf{Factuality}\\
    \midrule
    \textbf{English general model} & \textbf{GPT-4} & \textbf{4.97} & \textbf{4.33} & \textbf{3.06} \\
    \textbf{English financial model} & \textbf{FinLLaMA-8B} & 2.09 & 1.48 & 1.54 \\
    \textbf{Greek general model} & \textbf{Meltemi-7B} & \underline{3.99} & 1.49 & 1.60\\
    \textbf{Greek financial model} & \textbf{Plutus-8B} & 3.90 & \underline{3.51} & \underline{2.93}\\
  \bottomrule
\end{tabular}}
\end{table}

\begin{figure}[h]
  \centering
  \includegraphics[width=\linewidth]{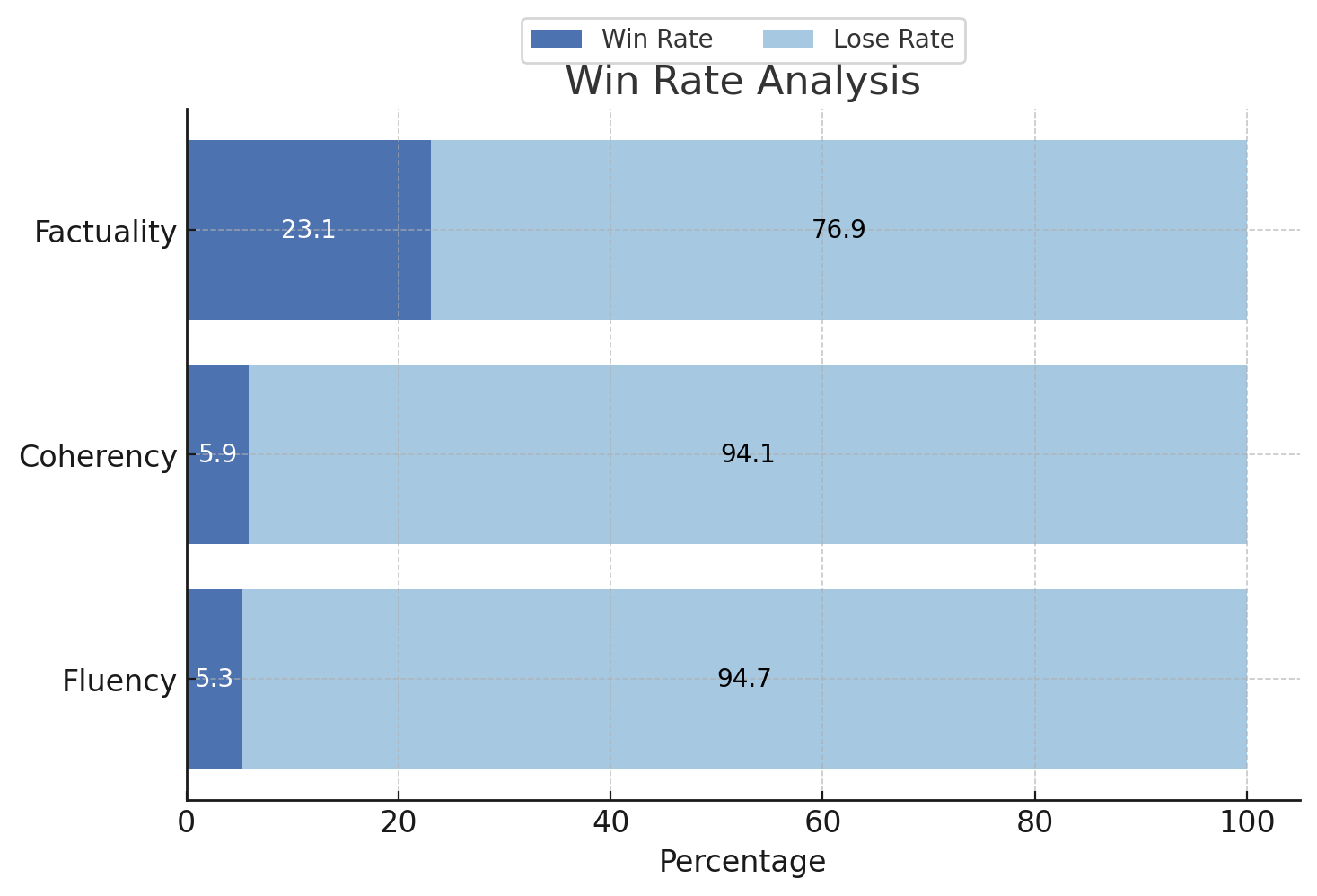}
  \caption{Comparison of model win rates in fluency, coherence, and factuality between Plutus-8B and GPT-4, evaluated on the GRFNS-2023 dataset within the Plutus-ben benchmark.}
    \label{fig:winrate}
\end{figure}

\subsection{Human Evaluation}

To gain deeper insights into models' performances on Greek language tasks, we conducted a standard human evaluation (Appendix~\ref{sec:humanevaluation_annotation}). 
The results presented in Table~\ref{tab:human_evaluation} show that while GPT-4 leads in fluency, domain-specific model like Plutus-8B excels other similar size models in Greek financial tasks, particularly in coherency and factuality. 
This highlights the need to enhance both linguistic and domain-specific capabilities to effectively adapt general-domain LLMs for specialized tasks.
Notably, Plutus-8B outperforms other models in terms of coherency (3.51) and factuality (2.93), showing significant improvements over other models of the same size. 
These results demonstrate that our designed tasks and curated high-quality dataset within the benchmark are effectiveness for improving models' factuality and coherency, which are challenging and important for financial tasks. The strong performance of Plutus-8B underscores the importance of language and domain-specific training, as models like FinLLaMA-8B — optimized for English financial data — struggle to adapt to Greek tasks.

Though Meltemi-7B, trained specifically for Greek general-purpose tasks, performs second in fluency with a score of 3.99 following GPT-4, its performance in coherency (1.49) and factuality (1.60) lags behind that of GPT-4 and Plutus-8B. This could be due to its Greek-specific training, which improves fluency but struggles with coherency and factuality in comparison to domain-optimized models. On the other hand, FinLLaMA-8B, trained on English financial data, performs poorly across all metrics, underscoring the challenges faced by English-centric models when applied to Greek.

We conducted a comparative analysis of Plutus-8B and GPT-4 using a win rate pairwise competition, focusing on their abilities in processing long-context data~\ref{fig:winrate}. Larger model sizes and advanced training methodologies are pivotal for effective long-context processing capabilities. In this regard, GPT-4 demonstrated superiority by outperforming Plutus-8B across nearly all samples in every dimension, attributed to its larger model size, extensive training data, and advanced long-context processing abilities.
The GRFNS-2023 dataset, which features the narrative sections of annual company reports as input, averaging around 60 pages and 31,500 words per document, poses a significant challenge for LLMs. Smaller models like Plutus-8B struggle to process such long-context inputs effectively, resulting in performance shortfalls across all dimensions. Nevertheless, Plutus-8B achieved a 23.1\% win rate in factuality rounds, with a performance score of 2.93 compared to GPT-4’s 3.06. This suggests that domain-specific training in Greek financial topics and language has equipped Plutus-8B with enhanced factual accuracy. 
The model benefits from instruction tuning that incorporates financial disambiguation terminology and numerical patterns unique to Greek texts, allowing Plutus-8B to grasp complex financial content more accurately and generate summaries with notable improvements in reliability and trustworthiness. This indicates the importance of targeted training in specific linguistic and domain contexts to enhance model performance.

% The win rates illustrated in Figure~\ref{fig:winrate} further delineate the comparative strengths of Plutus-8B and Meltemi-7B in Greek language tasks.
% Meltemi-7B demonstrates strong fluency, achieving a 66.6\% win rate and a fluency score of 3.99. Yet, its coherence (33.3\% win rate) and factuality (33.3\% win rate) fall short, indicating that while Greek-specific training bolsters fluency, it does not adequately address domain-specific reasoning or factual precision.
% Conversely, Plutus-8B, tailored for Greek financial tasks, exhibits robust domain understanding, with a 66.6\% win rate in both coherence and factuality. This aligns with its human evaluation scores: 3.51 for coherence and 2.93 for factuality, clearly surpassing Meltemi-7B. Although its fluency score of 3.90 is marginally lower than Meltemi-7B’s, the difference is minimal, suggesting that Plutus-8B retains competitive language proficiency while excelling in domain-specific knowledge.

Overall, Plutus-8B's domain-aware fine-tuning equips it to better navigate financial contexts—narrowing the gap with larger, general-purpose models like GPT-4. This highlights the critical role of combining linguistic and domain-specific training to enhance LLM performance in non-English, domain-focused tasks.

% In terms of \textbf{Fluency}, GPT-4 scores an impressive 4.98, categorized as “Excellent,” reflecting precise use of domain-specific terminology. Plutus-8B and Meltemi-7B achieve scores of 3.9 and 3.99, respectively, aligning with “Good,” producing generally fluent Greek text with minor grammatical errors. However, FinLLaMA-8B records a score of 2.1, marked as “Poor” due to a mix of Greek and English, which affects readability.

% Regarding \textbf{Coherence}, GPT-4 scores 4.34, indicating a “Good” level of logical structure and flow. Plutus-8B follows with a score of 3.51, categorized as “Okay,” showing basic coherence with some inconsistencies. In contrast, both FinLLaMA-8B and Meltemi-7B score around 1.5, rated as “Poor,” highlighting disorganized content and weak logical connections.

% For \textbf{Factuality}, GPT-4 achieves a score of 3.06, rated “Okay,” aligning closely with source content while exhibiting minor discrepancies. Plutus-8B scores 2.93, reflecting a similar level of factual alignment. Meanwhile, FinLLaMA-8B and Meltemi-7B, with scores around 1.5, fall into the “Poor” category due to significant factual inaccuracies.

\section{Conclusion}
In this study, we introduced \textbf{Plutus-ben}, the first comprehensive Greek financial evaluation benchmark, together with \textbf{Plutus-8B}, the inaugural Greek financial LLM, achieving SOTA performance on the Plutus-ben benchmark. These contributions address a notable gap, as there were previously no benchmarks or LLMs specifically tailored for Greek financial applications. \textbf{Plutus-ben} includes five essential NLP tasks—numeric and textual NER, QA, abstractive summarization, and topic classification—facilitating systematic and reproducible evaluations of LLMs in the Greek financial domain. Significantly, numeric and textual NER and QA are introduced in Greek for the first time. To support these tasks, we developed four high-quality datasets (GRFinNUM, GRFinNER, and GRFinQA), meticulously annotated by expert native Greek speakers with substantial financial and linguistic expertise, supplemented by two existing resources.
Our comprehensive evaluation of 22 LLMs, alongside a carefully designed human evaluation, highlights the intrinsic challenges of Greek financial NLP. These challenges arise from linguistic complexity, domain-specific terminology, gaps in financial reasoning, and the constraints of cross-lingual transfer. Notably, \textbf{Plutus-8B} demonstrated SOTA performance and impressive win rates in human evaluation, underscoring the importance of models trained with financial expertise and adapted to the nuances of Greek text.
By releasing Plutus-ben, and Plutus-8B, and all associated datasets, we aim to promote reproducible research and advance Greek financial NLP, fostering greater multilingual inclusivity in the financial sector and paving the way for further innovations and applications in this domain.

While this study provides valuable insights, our Plutus-ben benchmark is still limited in size and task types, and Plutus-8B's performance on the abstractive summarization task remains an area for improvement. In the future, we plan to expand Plutus-ben to include datasets spanning diverse tasks. Additionally, we aim to continually pretrain a Greek financial LLM with a larger model size to enhance efficiency and performance in the Greek financial context.

%%
%% The next two lines define the bibliography style to be used, and
%% the bibliography file.
\bibliographystyle{ACM-Reference-Format}
\bibliography{sample-base}

%%
%% If your work has an appendix, this is the place to put it.
\clearpage
\appendix
\onecolumn

\section{Open Greek Financial LLM Leaderboard}
\label{sec:leaderboard}

\begin{figure}[h]
  \centering
  \includegraphics[width=0.8\linewidth]{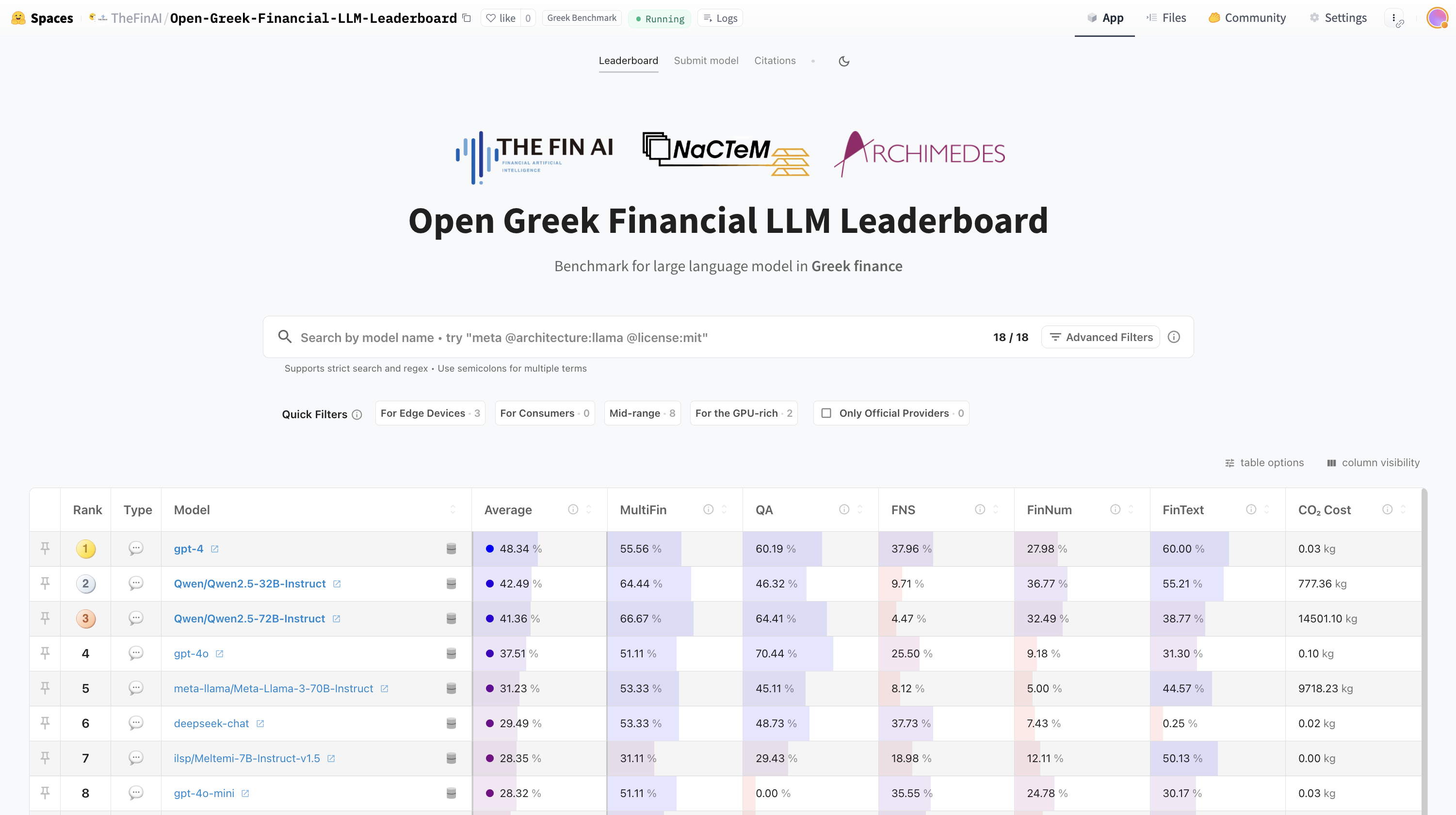}
  \caption{The Plutus-ben interface.}
    \label{fig:leaderboard}
\end{figure}

\section{Plutus-8B-instruct}
\label{sec:plutus-8B-instruct}

\begin{figure}[h]
  \centering
  \includegraphics[width=0.8\linewidth]{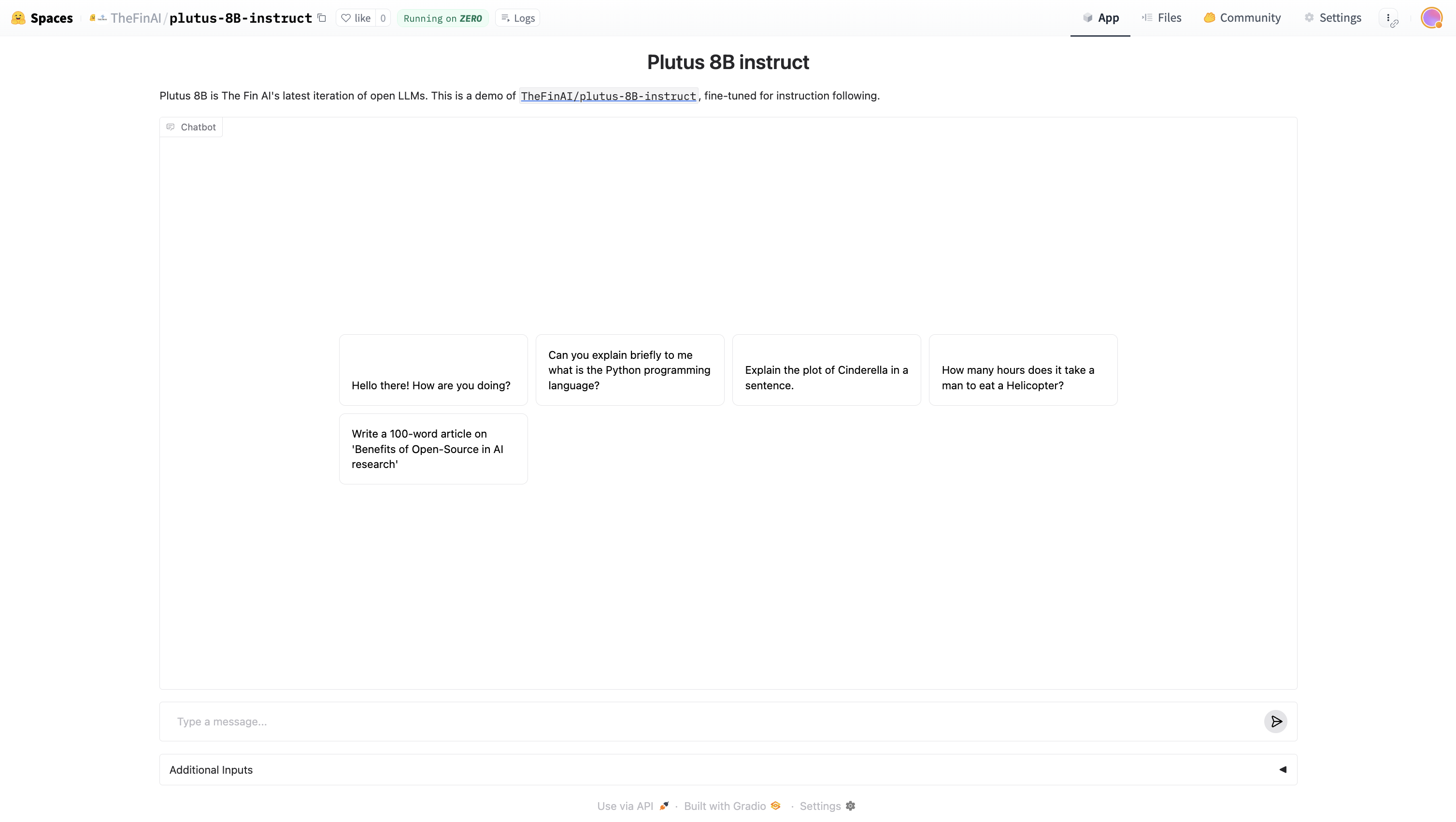}
  \caption{The Demo of Plutus-8B-instruct.}
    \label{fig:plutus}
\end{figure}

\newpage

\section{Dataset Curation and Conversion}
\label{sec:dataset}

\begin{table*}[!htbp]
\centering
\small
\caption{Datasets included in the Plutus-ben benchmark, presented in both the original Greek and their English translations.}
\label{tab:dataset}
% \scalebox{0.7}{
\resizebox{\textwidth}{!}{
    \begin{tabular}{@{}lccc@{}}
    \toprule
    \textbf{Dataset} & \textbf{Version} & \textbf{Input} & \textbf{Output} \\ 
    \midrule

    \multirow{2}{*}{GRFinNUM} 
    & Greek Original & 
    \makecell[l]{Σε επίπεδο ομίλου τα κέρδη ανα μετοχή είναι αυξημένα \\ + 10,11\% λόγω της επίδρασης λειτουργίας της Cosmokid AE \\ που ξεκίνησε ουσιαστικά το Β εξάμηνο του 2008.} & 
    \makecell[l]{10,11\%, ΠΟΣΟΣΤΑ \\ 2008, ΧΡΟΝΙΚΑ} \\ 
    & English Translation & 
    \makecell[l]{At the group level, earnings per share are increased \\ by +10.11\% due to the impact of Cosmokid AE's \\ operations, which started in the second half of 2008.} & 
    \makecell[l]{10.11\%, PERCENTAGE \\ 2008, TEMPORAL} \\ 
    \midrule

    \multirow{2}{*}{GRFinNER} 
    & Greek Original & 
    \makecell[l]{Στις 08.11.2019, η ΟΠΑΠ INVESTMENT LTD ήρθε σε συμφωνία \\ με την Εταιρεία για την πώληση του συνόλου των μετοχών \\ που κατέχει στην ΙΠΠΟΔΡΟΜΙΕΣ Α.Ε., έναντι συνολικού \\ τιμήματος € 10.411.} & 
    \makecell[l]{ΟΠΑΠ INVESTMENT LTD, ΟΡΓΑΝΙΣΜΟΣ \\ ΙΠΠΟΔΡΟΜΙΕΣ Α.Ε., ΟΡΓΑΝΙΣΜΟΣ} \\ 
    & English Translation & 
    \makecell[l]{On 08.11.2019, OPAP INVESTMENT LTD reached an agreement \\ with the Company for the sale of all \\ the shares it holds in HIPPODROMIES S.A., for €10,411.} & 
    \makecell[l]{OPAP INVESTMENT LTD, ORGANIZATION \\ HIPPODROMIES S.A., ORGANIZATION} \\ 
    \midrule

    \multirow{2}{*}{GRFinQA} 
    & Greek Original & 
    \makecell[l]{Βραχυχρονίως, μία αύξηση των δημοσίων δαπανών \\ Πιθανές απαντήσεις: \\ Α) αυξάνει το επίπεδο τιμών αλλά όχι το πραγματικό ΑΕΠ \\ Β) αυξάνει το πραγματικό ΑΕΠ αλλά όχι το επίπεδο τιμών \\ Γ) αυξάνει το πραγματικό ΑΕΠ και το επίπεδο τιμών \\ Δ) δεν αυξάνει ούτε το πραγματικό ΑΕΠ ούτε το επίπεδο τιμών} & 
    \makecell[l]{Γ} \\ 
    & English Translation & 
    \makecell[l]{In the short term, an increase in public spending \\ Possible answers: \\ A) Increases the price level but not real GDP \\ B) Increases real GDP but not the price level \\ C) Increases both real GDP and the price level \\ D) Increases neither real GDP nor the price level} & 
    \makecell[l]{C} \\ 
    \midrule

    % \multirow{2}{*}{GRFinSUM} 
    % & Greek Original & - & - \\ 
    % & English Translation & - & - \\ 
    % \midrule

    \multirow{2}{*}{GRFNS-2023} 
    & Greek Original & \makecell[l]{Τα μέλη του Διοικητικού Συμβουλίου της \\ ΚΑΠΝΟΒΙΟΜΗΧΑΝ (...TRUNCATED)} & 
    \makecell[l]{Ετήσια Οικονομική Έκθεση της Χρήσης \\ ΔΩΔΕΚΑΜΗΝΗ ΠΕΡΙΟ (...TRUNCATED)} \\ 
    & English Translation & \makecell[l]{The members of the Board of Directors of TOBACCO \\ INDUSTRY (...TRUNCATED)} & 
    \makecell[l]{Annual Financial Report for the TWELVE-MONTH \\ PERIOD (...TRUNCATED)} \\ 
    \midrule

    \multirow{2}{*}{GRMultiFin} 
    & Greek Original & 
    \makecell[l]{Αναστολή συμβάσεων εργασίας Αυγούστου} & 
    \makecell[l]{Επιχειρήσεις \& Διοίκηση} \\ 
    & English Translation & 
    \makecell[l]{Suspension of employment contracts in August} & 
    \makecell[l]{Business \& Administration} \\ 

    \bottomrule
    \end{tabular}
}
\end{table*}

\begin{table*}[!htbp]
\centering
\small
\caption{Conversion prompts for instruction data, presented with original Greek prompts alongside their English translations.}
\label{tab:prompt}
\resizebox{\textwidth}{!}{
    \begin{tabular}{@{}lcc@{}}
    \toprule
    \textbf{Dataset} 
    & \textbf{Original Greek Prompt} 
    & \textbf{English Translated Prompt} \\ 
    \midrule
    GRFinNUM & \makecell{Στις παρακάτω προτάσεις που προέρχονται από οικονομικές εκθέσεις \\ ελληνικών εταιρειών, αναγνώρισε αριθμητικές οντότητες που \\ ανήκουν στις εξής κατηγορίες: χρηματικά ποσά (ΧΡΗΜΑΤΑ), \\ ποσοστά (ΠΟΣΟΣΤΑ), χρονικές τιμές (ΧΡΟΝΙΚΑ), ποσότητες (ΠΟΣΟΤΗΤΕΣ) \\ και άλλες αριθμητικές τιμές (ΑΛΛΑ). Η απαιτούμενη μορφή απάντησης \\
    είναι 'όνομα οντότητας, τύπος οντότητας'. Κείμενο: \{Input\} Απάντηση:} & \makecell{In the following sentences which originate from Greek Company filings, \\recognize the numeric entities which correspond to the following categories: \\monetary values (MONETARY), percentages (PERCENTAGES), temporal \\values (TEMPORAL), quantities (QUANTITIES) and other numeric values \\(OTHER). The required answer format is: ``entity name, entity type''. Text: \\\{Input\} Answer:} \\ 

    \midrule
    GRFinNER & \makecell{Στις παρακάτω προτάσεις που προέρχονται από οικονομικές εκθέσεις \\ ελληνικών εταιρειών, αναγνώρισε τις οντότητες που αντιπροσωπεύουν \\ ένα πρόσωπο (ΠΡΟΣΩΠΟ), έναν οργανισμό (ΟΡΓΑΝΙΣΜΟΣ) ή μία \\ τοποθεσία (ΤΟΠΟΘΕΣΙΑ). Η απαιτούμενη μορφή είναι: \\ 'όνομα οντότητας, τύπος οντότητας'. Κείμενο: \{Input\} Απάντηση:} & \makecell{In the following sentences which originate from Greek Company filings, \\recognize the entities which correspond to a person ("Person"), an \\organization ("Organisation") or a location ("Location"). The required \\answer format is: ``entity name, entity type''. Text: \{Input\} Answer:} \\ 

    \midrule
    GRFinQA & \makecell{Διάβασε προσεκτικά την παρακάτω ερώτηση και τις πιθανές απαντήσεις. \\ Επίλεξε το γράμμα που αντιστοιχεί στη σωστή απάντηση.\\ Ερώτηση: \{Input\} Απάντηση:} & \makecell{Read the following question and the possible answers carefully. Choose the \\letter which corresponds to the correct answer. Question: \{Input\} Answer:} \\ 

    % \midrule
    % GRFinSUM & - & - \\ 

    \midrule
    GRFNS-2023 & \makecell{Σε παρακαλώ διάβασε το παρακάτω κείμενο και συνόψισε το σύντομα και με ακρίβεια. \\ \{Input\}} & \makecell{Please read the following text and summarize it briefly and accurately. \\\{Input\}} \\

    \midrule
    GRMultiFin & \makecell{Διάβασε το κείμενο προσεκτικά και επέλεξε την σωστή κατηγοριά για το \\ κείμενο από τις κατηγορίες Φορολογία \& Λογιστική, Επιχειρήσεις \& Διοίκηση, \\ Οικονομικά, Βιομηχανία, Τεχνολογία, Κυβέρνηση \& Έλεγχοι. \\ Κείμενο: \{Input\} Απάντηση:} & \makecell{Read the text carefully and choose the correct category for the text from the \\categories ``Tax \& Accounting'', ``Business \& Management'', ``Finance'', \\``Industry'', ``Technology'', ``Government \& Controls''. Text: \{Input\} Answer:} \\ 

    \bottomrule
    \end{tabular}
    }
\end{table*}

\section{GRFinNUM Annotation Guideline}
\label{sec:numner_annotation}

To ensure consistent annotation of numerical entities in financial texts, we define the following annotation guidelines.

\subsection{Entity Categories}

We annotate five types of numerical entities:

\begin{itemize}
    \item \textbf{Monetary}
    \item \textbf{Percentage}
    \item \textbf{Temporal}
    \item \textbf{Quantity}
    \item \textbf{Others}
\end{itemize}

\subsection{General Annotation Rules}

\begin{enumerate}
    \item \textbf{Only numbers are annotated}: Include only numerical digits, decimal points (``.''), and the percent sign (``\%'').
    \item \textbf{Decimal delimiter exclusion}: When a decimal point is used as a delimiter (e.g., 2024.11.26), annotate each component separately as 2024, 11, and 26.
    \item \textbf{Exclusion of textual numbers}: Text-based numbers (e.g., two weeks) are excluded, but numeric equivalents (e.g., 2 weeks) are included.
    \item \textbf{Exclusion of non-numeric symbols}: Symbols such as ``\$'' are not included.
\end{enumerate}

\subsection{Specific Entity Annotation Rules}

\subsubsection{Monetary}
Numbers related to money, including explicit currencies or monetary values.

\begin{itemize}
    \item Include: The numeric value in ``\$50'' and ``100 euros'' → annotate as ``50'' and ``100''.
\end{itemize}

\subsubsection{Percentage}
Numbers representing percentages, ``\%'' symbol as part of the percentage.

\begin{itemize}
    \item Include: ``45\%'', ``0.5\%''.
\end{itemize}

\subsubsection{Temporal}
Numbers related to time, such as years, dates, and durations.

\begin{itemize}
    \item Include:  only numbers in ``2024'', ``12.25'', ``12/25'', ``2 weeks'', ``1 year'' and ``3 hours'' should be included.
    \item Exclude: Words such as ``two weeks'', where the number is not explicitly written in numeric form.
\end{itemize}

\subsubsection{Quantity}
Numbers representing measurable or countable quantities, excluding monetary values.

\begin{itemize}
    \item Include: only numbers in ``5 items'' and ``100 shares''.
\end{itemize}

\subsubsection{Others}
Numbers that do not fit into the above categories, such as identifiers, version numbers, numerical codes, or numeric positions.

\begin{itemize}
    \item Include: only ``3'' in ``3rd place'', ``2'' and ``1'' in ``v2.1'', and ``202'' in ``model 202''.
    \item Exclude: ``second investor'' (textual ordinal numbers).
\end{itemize}

\subsection{Annotation Examples}

\begin{table}[h]
\centering
\begin{tabular}{p{5cm} p{3cm}}
\hline
\textbf{Text} & \textbf{Annotated Entity} \\
\hline
``\$50 was paid.'' & `50' (Monetary) \\
``45\% of users agreed.'' & `45\%' (Percentage) \\
``The event happened in 2024.'' & `2024' (Temporal) \\
``5 items were sold.'' & `5' (Quantity) \\
``Version v2.1 is released.'' & `2', `1' (Others) \\
\hline
\end{tabular}
\caption{Examples of annotated numerical entities.}
\label{tab:num_annot_examples}
\end{table}

\section{GRFinNER Annotation Guideline} \label{sec:textner_annotation}

To ensure consistent annotation of named entities in financial texts, we define the following annotation guidelines.

\subsection{Entity Categories}

We annotate three types of named entities:

\begin{itemize} 
    \item \textbf{Person} 
    \item \textbf{Location} 
    \item \textbf{Organization} 
\end{itemize}

\subsection{General Annotation Rules}

\begin{enumerate} 
    \item \textbf{Abbreviations}: Annotate them together if they appear together; otherwise, annotate them as two entities.
    
    \begin{itemize} 
        \item Include: ``World Health Organization (WHO)'' as one span.
    \end{itemize}

    \item \textbf{Ambiguous Terms}: Resolve ambiguity using context.
    \begin{itemize}
        \item Include: ``Amazon'' as a company.
        \item Exclude: ``Amazon'' as a river.
    \end{itemize}

    \item \textbf{General Terms Exclusion}: Exclude generic terms.
    \begin{itemize}
        \item Exclude: ``the professor'', ``downtown'', ``north'', ``the team''.
    \end{itemize}

    \item \textbf{Definite Articles}: Exclude ``the'' from entity spans.
    \begin{itemize}
        \item Exclude: ``the'' in ``the WHO''.
    \end{itemize}

    \item \textbf{Consecutive Entities}: When two entities are consecutive, annotate them separately except postal addresses.
    \begin{itemize}
        \item Include separately: ``London'' and ``United Kingdom'' in ``London United Kingdom''.
        \item Include separately: `street Egnatias 127'' and ``Thessaloniki'' in ``street Egnatias 127 in Thessaloniki (Postal Code 54 635)''.
        \item Include separately: ``Acharnes Attica'' and ``Parnithos Avenue'' in ``municipality of Acharnes Attica, 15 km Parnithos Avenue''.
        \item Include as one span: ``5900 Penn Avenue, Pittsburgh''.
    \end{itemize}

\end{enumerate}

\subsection{Specific Entity Annotation Rules}

\subsubsection{Person} Names of individual people. Include real people, fictional characters, and usernames. Exclude animal names. Exclude titles that are not part of the legal name.

\begin{itemize} 
    \item Include: ``Marie Curie'', ``George Demetriou of Konstantinos''. 
    \item Include only `John'' in `Dr. John''. 
    \item Exclude: ``the professor''.
\end{itemize}

\subsubsection{Location} Names of geographical places, such as cities, countries, natural landmarks, and fictional locations.

\begin{itemize} 
    \item Include: `Paris'', `Mount Everest''. 
    \item Exclude: `downtown'', `north''. 
\end{itemize}

\subsubsection{Organization} Names of companies, institutions, and formal groups. Including words like ``company'', ``association'', ``Inc.'', ``Co.'', and ``Ltd.''.

\begin{itemize} 
    \item Include: ``World Health Organization'', ``Tesla Inc.'', ``WHO'', ``OPAP Association''. 
    \item Exclude: ``the team''. 
\end{itemize}

\subsection{Special Cases}

\begin{enumerate} 
    \item \textbf{Organizations with Location Names}: If the location refers to a specific organization, annotate both; otherwise, only annotate the location. 
    \begin{itemize} 
        \item Include: Only `Cypriot'' in `the Cypriot company''.
    \end{itemize}

    \item \textbf{Organizations Representing Administrative Units or Sports Teams}: Annotate as \textbf{Organization}.
    \begin{itemize}
        \item Include: ``Baltimore'' and ``Indianapolis'' in ``Baltimore lost to Indianapolis last weekend'' as Organizations.
    \end{itemize}

\end{enumerate}

\section{GRFinSUM Annotation Guideline}
\label{sec:sum_annotation}

To ensure consistency and accuracy in extractive summarization, we established a set of annotation guidelines for identifying and selecting relevant textual segments. Our approach focuses on extracting key financial metrics while excluding extra narrative content which elaborates upon other sentences. All annotations were conducted at the sentence level, ensuring that complete sentences are selected. We also include the final punctuation mark for each sentence. The following rules were applied during the annotation process:

The text contains a substantial amount of financial data; however, not all financial metrics are included in the annotations. Our selection criteria prioritize core earnings and expense-related metrics, while excluding explanatory narratives, interpretations, or alternative financial indicators.

In principle metrics can be seperated into two categories, earnings related metrics and expense related metrics. For each category we focus on the following specific indicators:

Earnings Metrics:

We annotate sentences which contain information about Pre-tax earnings, 
%(Κέρδη προ φόρων) 
After-tax earnings (Net profit), 
%(Κέρδη μετά φόρων)
Revenue/Turnover, 
%(Κύκλος εργασιών)
Profit margin, 
%(Περιθώριο κέρδους)

However, we do not annotate all earnings related metrics. We exclude metrics such as EBIT (Earnings Before Interest and Taxes) and EBITDA (Earnings Before Interest, Taxes, Depreciation, and Amortization) from our annotations. While these are commonly used financial indicators, they are considered alternative or non-GAAP metrics and are not universally standardized under frameworks such as IFRS.

Expense Metrics:

We annotate expenses, operating expenses and total expenses, 
%(Έξοδα, λειτουργικά έξοδα, σύνολο εξόδων)
prioritizing the most generic form of the expense metric in each instance.
When financial reports provide a breakdown of expenses by specific projects or operational segments, only the aggregate expense value is included unless the breakdown is contained within the same sentence. Complementary explanations regarding how specific projects contribute to overall costs are excluded from the annotation.

\section{Human Evaluation Annotation Guideline}
\label{sec:humanevaluation_annotation}

To ensure consistent annotation of summarization quality in financial texts, we define the following annotation guidelines.

\subsection{Evaluation Categories}

We evaluate summaries based on three criteria:

\begin{itemize} 
    \item \textbf{Language Appropriate Fluency} 
    \item \textbf{Coherence} 
    \item \textbf{Factuality} 
\end{itemize}

\subsection{General Annotation Rules}

\begin{enumerate} 
    \item \textbf{Language Appropriate Fluency (Fluency)}: Measures how well the summary aligns with the expected language fluency and domain-specific terminology.
    \begin{itemize} 
        \item 1 (Bad): Response is entirely in the wrong language (e.g., English instead of Greek).
        \item 2 (Poor): Response is a mixture of English and Greek.
        \item 3 (Okay): Response is fully in Greek but contains grammatical or lexical errors or repetition.
        \item 4 (Good): Response is entirely in fluent Greek without grammatical or lexical errors or repetition.
        \item 5 (Excellent): Response is entirely in fluent Greek with appropriate domain-specific terminology.
    \end{itemize}

    \item \textbf{Coherence}: Evaluates the logical progression and structure of ideas in the text.
    \begin{itemize} 
        \item 1 (Bad): The text is disorganized, with sentences or paragraphs lacking logical flow.
        \item 2 (Poor): The text attempts structure but has logical leaps, disjoint ideas, and is confusing.
        \item 3 (Okay): The text is mostly coherent, with a general structure and minor logical errors or awkward transitions.
        \item 4 (Good): The text flows well, with clear progression and only minor errors.
        \item 5 (Excellent): The text flows naturally and consistently, with smooth transitions between ideas.
    \end{itemize}

    \item \textbf{Factuality}: Evaluates whether the summary is factually consistent with the original content.
    \begin{itemize} 
        \item 1 (Bad): Multiple factual inaccuracies, such as misrepresented company names, locations, or numerical data.
        \item 2 (Poor): Some factual errors with key points missing or distorted.
        \item 3 (Okay): Fairly accurate, with only minor omissions or discrepancies.
        \item 4 (Good): Accurate, with only a few minor omissions or discrepancies.
        \item 5 (Excellent): Entirely accurate, with all facts presented as found in the source document.
    \end{itemize}
\end{enumerate}

\section{Annotator Demography}
\label{sec:annotator}
% Three native Greek speakers with extensive experience living in Greece contributed to building our first Greek financial benchmark. They all hold advanced degrees in relevant fields and possess experience in institutes specializing in economics and business. 

Our benchmark construction relies on the expertise of a team of highly qualified annotators, who are native Greek speakers with diverse backgrounds in computer science, mathematics, statistics, and finance. Their combined knowledge ensures the high-quality annotation of financial texts, contributing to the robustness and reliability of our dataset.

One annotator, currently pursuing a Ph.D. in Computer Science at a leading Greek university, has a strong foundation in both mathematics and statistics, complemented by industry experience as a credit risk analyst. This background provides valuable information on financial knowledge, risk assessment, and statistical modeling, which are essential to annotate our benchmark dataset.

Another annotator, a Ph.D. student in Computer Science at a major UK institution, holds an Integrated Master's degree in Electrical and Computer Engineering. Their expertise in computer science enhances the annotation process by ensuring precision and alignment with modern NLP techniques.

The team is further strengthened by a postdoctoral researcher with an interdisciplinary background spanning electrical and computer engineering, computer science, and mathematics. Having obtained a Ph.D. from a prestigious U.S. university, this annotator brings extensive research experience and a deep understanding of theoretical and applied aspects of financial computing, making them instrumental in refining annotation guidelines and resolving complex cases.

The collective expertise of our annotators is critical to the development of our Greek financial benchmark. Their deep familiarity with the Greek financial ecosystem, combined with strong computational and analytical skills, ensures that our dataset accurately reflects domain-specific nuances while maintaining linguistic and terminological precision. By leveraging their diverse backgrounds, we are able to construct a high-quality resource that will serve as a foundation for advancing NLP research in financial applications.

% Triantafillos Papadopoulos
% Greek Ph.D. student in Computer Science at AUEB with a Master's degree in Statistics from AUEB and a Bachelor's degree in Mathematics from NKUA, with one-year work experience as a credit risk analyst.

% Polydoros Giannouris
% Greek Ph.D. student in Computer Science at the University of Manchester with a Integrated Master's degree in electrical and computer engineering from AUTH.

% Efstathia Soufleri
% Greek Postdoctoral Researcher with a Ph.D. in Electrical and Computer Engineering from Purdue University (USA), a Master's degree in Computer Science from the University of Thessaly, and a Bachelor's degree in Mathematics from the National and Kapodistrian University of Athens.

\section{Annotation Process}
\label{sec:labelstudio}

\begin{figure}[h]
  \centering
  \includegraphics[width=0.95\linewidth]{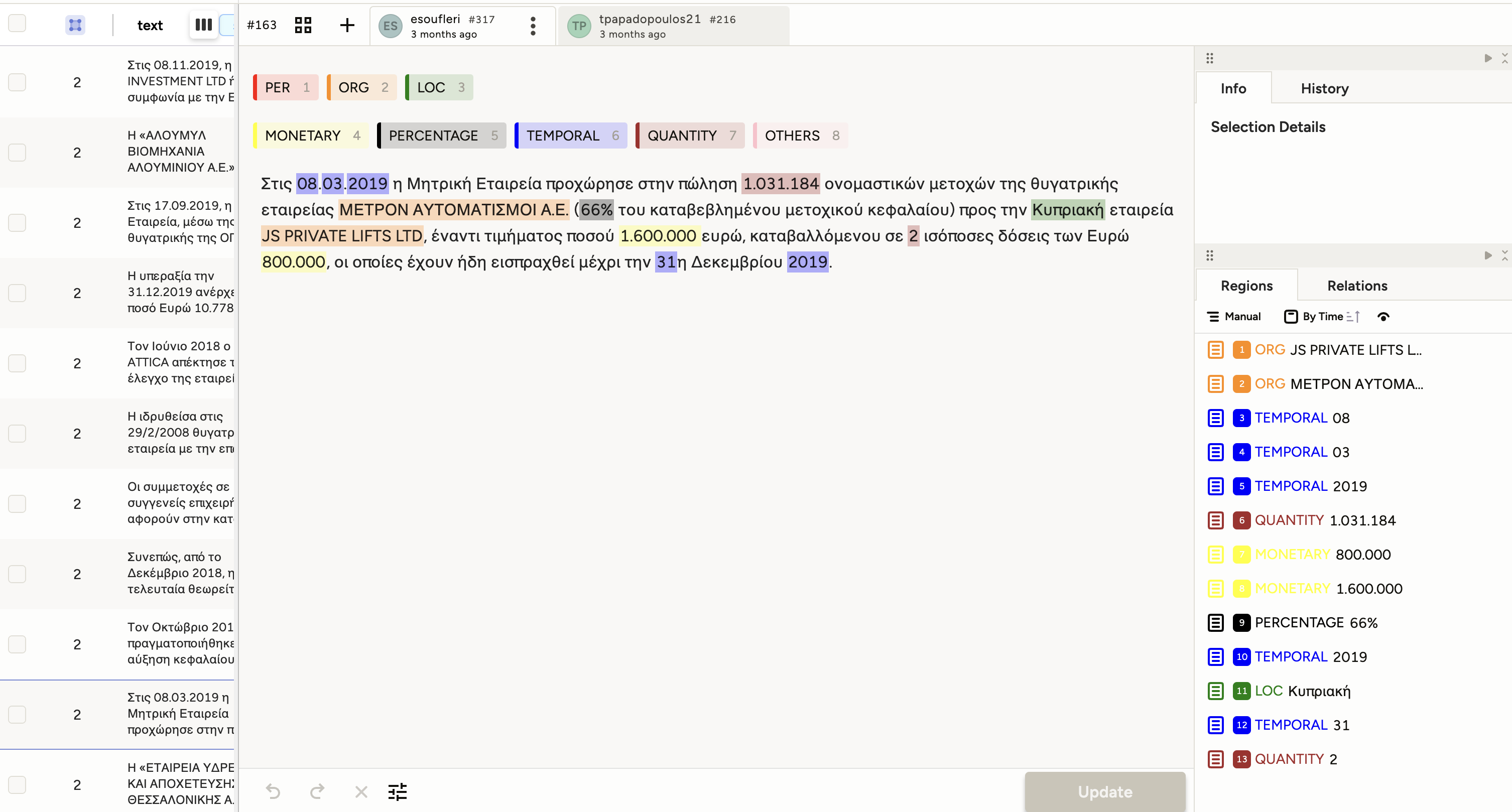}
  \caption{The Label Studio interface of the NER annotation process.}
    \label{fig:ner}
\end{figure}
\begin{figure}[h]
  \centering
  \includegraphics[width=0.95\linewidth]{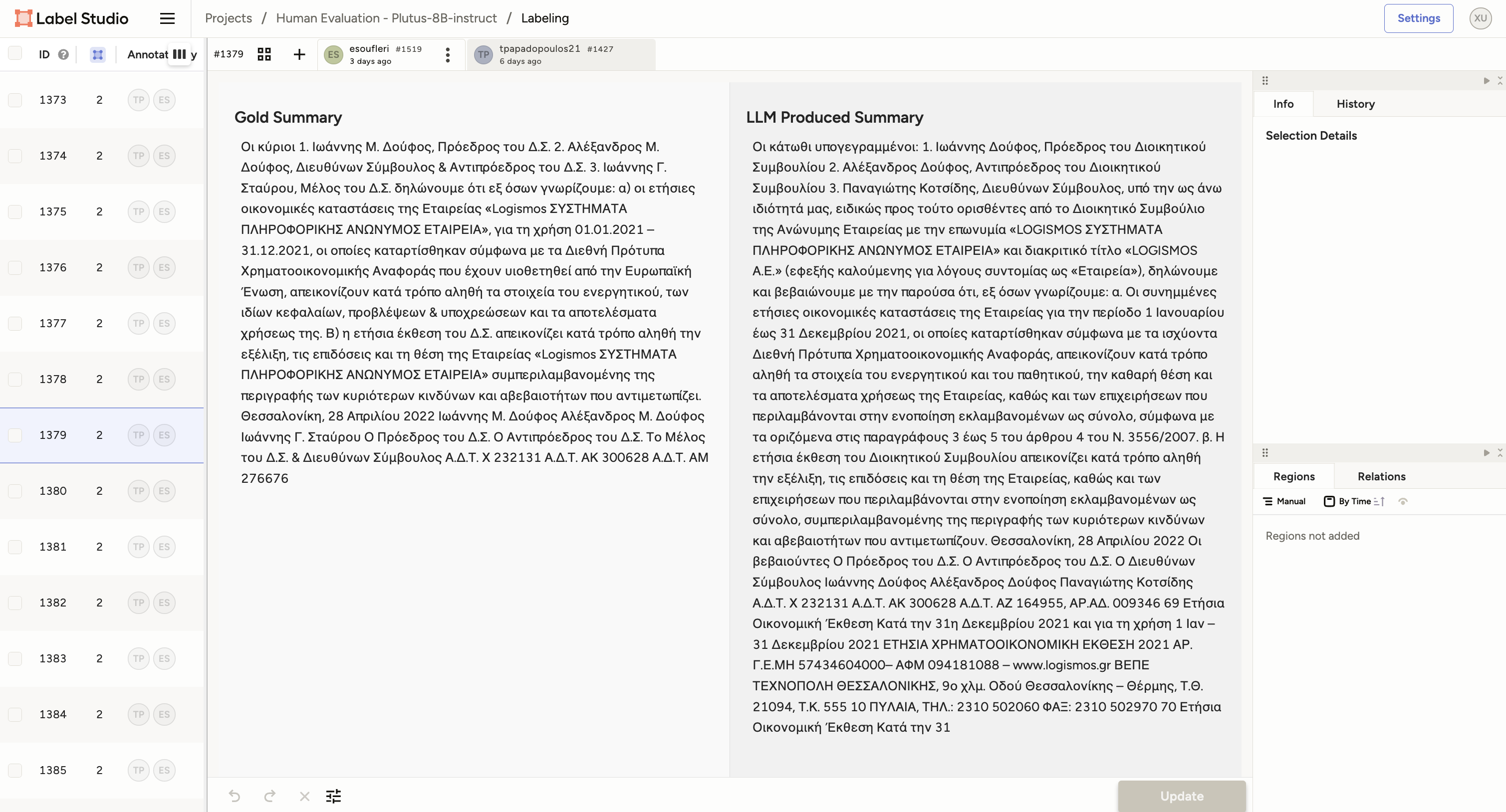}
  \caption{The Label Studio interface of the human evaluation process.}
    \label{fig:humanevaluation}
\end{figure}

\newpage

% Pre-annotation
% Annotation round 
\section{Model Evaluation}
\label{sec:model_evaluation}

The models are categorized as follows:
\textbf{GPT series}: GPT-3.5-Turbo~\cite{brown2020languagemodelsfewshotlearners}, GPT-4o-Mini~\cite{openai2024gpt4technicalreport}, GPT-4o~\cite{hurst2024gpt}, and GPT-4~\cite{achiam2023gpt}.
\textbf{LLaMA series}: LLaMA-3.2-1B~\citep{dubey2024llama}, LLaMA-3-8B~\citep{dubey2024llama}, LLaMA-3.1-8B~\citep{dubey2024llama}, and LLaMA-3-70B~\citep{dubey2024llama}.
\textbf{Qwen series}: Qwen2.5-1.5B~\cite{qwen2.5}, Qwen2.5-7B~\cite{qwen2.5}, Qwen2.5-32B~\cite{qwen2.5}, and Qwen2.5-72B~\cite{qwen2.5}.
\textbf{Gemma series}: Gemma-2-2B~\cite{team2024gemma}, Gemma-2-9B~\cite{team2024gemma}, and Gemma-2-27B~\cite{team2024gemma}.
\textbf{Mistral-7B~\cite{Mistral7b}}: Implemented as the foundation of the Meltemi-7B, Mistral-7B serves as a contrasting baseline model.
\textbf{Finma-7B~\cite{xie2023pixiu}}: Finma-7B is fine-tuned with large-scale multi-task instruction data, enhancing its utility in financial-specific task engagement.
\textbf{FinLLaMA-8B~\cite{xie2024open}}: FinLLaMA-8B is instruction fine-tuned with 573K financial instructions, excelling in navigating contextually complex financial discourse.
\textbf{Meltemi-7B~\cite{voukoutis2024meltemiopenlargelanguage}}: Built on Mistral-7B with continual pretraining in Greek and English, Meltemi-7B exhibits strong linguistic capabilities, though its financial methodology skills are untested.
\textbf{Llama-Krikri-8B\footnote{\url{https://huggingface.co/ilsp/Llama-Krikri-8B-Base}}}: Based on Llama-3.1-8B with continual pretraining with Greek, English, and math and coding data, Llama-Krikri-8B demonstrates robust linguistic abilities but lacks validation in financial domains.
\textbf{Plutus-8B}: Derived from Llama-Krikri-8B, Plutus-8B is instruction fine-tuned using Greek financial data, enhancing its capacity for specialized financial reasoning.

\section{Evaluation Metrics}

The Entity F1 is the harmonic mean of Precision and Recall, calculated as follows.
\begin{equation}
    P_{entity} = \frac{TP}{TP+FP}
\end{equation}

\begin{equation}
    R_{entity} = \frac{TP}{TP+FN}
\end{equation}

\begin{equation}
    \text{Entity F1} = 2 \times \frac{P_{entity} \times R_{entity}}{P_{entity} + R_{entity}}
\end{equation}
where $P_{entity}$ and $R_{entity}$ denote the Precision and Recall of entity prediction, respectively. $TP$ (True Positive) represents the number of actual entities correctly identified. In contrast, $FP$ (False Positive) refers to the number of non-entities incorrectly predicted as entities. $FN$ (False Negative) denotes the number of entities that were not correctly predicted.

Accuracy $\mathrm{Acc}$ measures the proportion of correct predictions made by the model and is defined as follows.

\begin{equation}
    \mathrm{Acc} = \frac{\text{Number of correct predictions}}{\text{Total number of Predictions}}
\end{equation}

Rouge-1 is primarily used to compute the unigram-level (word-level) overlap between the generated summary and the reference summary, and is defined as follows:

\begin{equation}
    P_{rouge1} = \frac{\text{Number of overlapping unigrams in generated and reference summary}}{\text{Total unigrams in generated summary}}
\end{equation}

\begin{equation}
    R_{rouge1} = \frac{\text{Number of overlapping unigrams in generated and reference summary}}{\text{Total unigrams in reference summary}}
\end{equation}

\begin{equation}
    \text{Rouge-1 F1}=2\times \frac{P_{rouge1} \times R_{rouge1}}{P_{rouge1} + R_{rouge1}}
\end{equation}
where $P_{rouge1}$ and $R_{rouge1}$ denote the Precision and Recall of Rouge-1, respectively. Rouge-1 F1 is the final Rouge-1 score that calculates the unigram (single-word) matches without considering word order. 

\section{Dataset Quality Validation}
\label{sec:agreement}

The F1-score, Cohen’s Kappa, and Krippendorff's alpha were calculated to measure the agreement of annotators for data quality control purposes.

The F1-score is a performance metric for classification models that combines Precision and Recall using their harmonic mean as shown in the equation~(\ref{eq:f1}).

\begin{equation}
    F1-scores = \frac{2 \cdot Precision\cdot Recall}{Precision + Recall}
    \label{eq:f1}
\end{equation}
where \textit{Precision} measures how many of the samples predicted as positive are actually positive; \textit{Recall} measures the proportion of actual positive samples that the model correctly identifies.

Cohen’s Kappa measures the agreement between two annotators on a classification task, accounting for the possibility of random agreement, as shown in equation~(\ref{eq:kappa}).

\begin{equation}
    \kappa = \frac{P_o - P_e}{1- P_e}
    \label{eq:kappa}
\end{equation}
where $P_o$ means the observed agreement and $P_e$ is the expected agreement.

Krippendorff’s alpha is a general measure of inter-rater reliability applicable to categorical, ordinal, interval, or ratio data, as shown in equation~(\ref{eq:alpha}).

\begin{equation}
    \alpha = 1 - \frac{D_o}{D_e}
    \label{eq:alpha}
\end{equation}
where $D_o$ is the total disagreement observed among annotators, and $D_e$ is the total disagreement expected by chance.

% \section{Limitations}
% While this study offers valuable insights, it is important to acknowledge the following limitations:
% (1) \textbf{Parameter Restriction}: Plutus-8B is dcurrently limited to a size of 8B parameters, and future work should explore both smaller models for efficiency and larger models for enhanced performance. 
% (2) \textbf{Limited Evaluation Benchmark}: The datasets available in Plutus-ben are limited in size, which may impede the model's ability to understand financial contexts comprehensively and generalize effectively across diverse scenarios. Plutus-8B exhibits varied performance on Plutus-ben, particularly struggling with summarizing long-form financial documents.
% (3) \textbf{Limited Application Scope}: The design and instructional approach of Plutus-8B may constrain its utility across different bilingual contexts. This specific tailoring could limit its generalizability to other linguistic or cultural scenarios.
% (4) \textbf{Ethical and Practical Concerns}: We must consider the potential for negative outcomes, such as disseminating inaccurate financial information or improper market influence. Therefore, we recommend utilizing Plutus-8B primarily for scholarly research, mindful of these ethical aspects.

\section{Ethical Statement}
The authors take full responsibility for the development and dissemination of Plutus-ben and Plutus-8B, ensuring that all raw data used are publicly available, devoid of personal information, and conform to established ethical guidelines. The data are shared under the MIT license, requiring users to adhere to its terms. This manuscript, including large language models, source codes, and datasets, is intended for academic and educational purposes only and is not a substitute for professional advice. While efforts have been made to ensure its accuracy, the authors and their institutions disclaim liability for any outcomes arising from its use. Users agree to take responsibility for ethical and lawful use and to indemnify the authors and their affiliates against any claims or damages resulting from reliance on this Material.

\end{document}